%% file: thesis.tex
\documentclass[cornellheadings,smallerheadings]{fiu}
\usepackage[left=1.5in, right=1.0in, bottom=1.25in, top=1.0in]{geometry}
\usepackage{graphicx}
\usepackage[linesnumbered, ruled,vlined]{algorithm2e}
\usepackage[normalem]{ulem}
\useunder{\uline}{\ul}{}
\usepackage{amsmath}
\usepackage[table,xcdraw]{xcolor}
\usepackage{listings}
\usepackage{adjustbox}
\usepackage{array}
\usepackage{appendix}
\usepackage{hyperref}
\DeclareGraphicsExtensions{.pdf,.jpeg,.png}


\title{A Bayesian Programming Approach to Car-following Model Calibration and Validation using Limited Data}
\author{Franklin Abodo}
\conferraldate{July}{2022}
\defensedate{June 24, 2022}
\advisor{Leonardo Bobadilla}
\memberone{Patricia McDermott-Wells}
\membertwo{Mark Finlayson}
\degreefield{Computer Science}
\college{College of Engineering and Computing}
\collegedean{Dean John L. Volakis}
\gradschooldean{Andrés G. Gil}

\begin{document} 

\setcounter{page}{1}
\pagenumbering{roman}
\pagestyle{plain}

\include{prologue}

\normalspacing
\setcounter{page}{1}
\pagenumbering{arabic}
\pagestyle{cornell}

\include{chap1}
\include{chap2}
\include{chap3}
\include{chap4}
\include{chap5}
\include{chap6}
\include{chap7}

\include{chap8}
\include{epilogue}
\include{appendix}

\end{document}

%% file: prologue.tex

\maketitle

\makeapproval{3}

\makecopyright

\begin{dedication}

This thesis is dedicated firstly to my mother, Judith Edwards, who has sacrificed selflessly day in and day out to maximize my quality of life over the course of my entire life, and whose loving example has always served to remind me that it is possible for us to be decent, just, and kind in an indecent, unjust, and unkind world as long as we follow our conscience and stay true to ourselves. This thesis is dedicated secondly to my long late father, Peter Abodo, who established starting early in my childhood that education would be the key path to a prosperous life, and who left me with a wealth of love-filled memories that remind me that happiness is an achievable state of existence.

\end{dedication}

\begin{acknowledgments}

I want to thank Leonardo Bobadilla immensely for being consistently and persistently good-willed in his effort to guide me through this project and, more broadly, through my time at FIU, and for being most generous with his time for the many years that he has been my research advisor.

I also want to thank Patricia McDermott-Wells for being a mentor to and advocate for me during my undergraduate tenure at FIU, and for always encouraging me to keep aspiring to greatness in spite of self-doubts. I am particularly grateful for the debates and respectful disagreements that we shared about gender and race in our society. Having those memories as a reminder that we can respect and even be friends with people who have different points of view in life helps me to remain open minded during this polarized period of our country's history.

Many thanks as well to Mark Finlayson for always seeking over-achievement, whether as a professor or as a friend, the results of which benefit everyone around you, and of course for serving on my thesis committee.

\end{acknowledgments}

\begin{abstract}
\input{abstract}
\end{abstract}

\contentspage

\tablelistpage

\figurelistpage

%% file: abstract.tex

Traffic simulation software is used by transportation researchers and engineers to design and evaluate changes to roadway networks. Underlying these simulators are mathematical models of microscopic driver behavior from which macroscopic measures of flow and congestion can be recovered. Many models are intended to apply to only a subset of possible traffic scenarios and roadway configurations, while others do not have any explicit constraint on their applicability. Work zones on highways are one scenario for which no model invented to date has been shown to accurately reproduce realistic driving behavior. This makes it difficult to optimize for safety and other metrics when designing a work zone. 

The Federal Highway Administration (FHWA) has commissioned the Volpe National Transportation Systems Center (Volpe) to develop a new car-following model, the Work Zone Driver Model (WZDM), for use in microscopic simulators that captures and reproduces driver behavior equally well within and outside of work zones. Volpe also performed a naturalistic driving study (NDS) to collect telematics data from vehicles driven on highways and urban roads that included work zones for use in model calibration. The data variables are relevant to the car-following model’s prediction task. 

During model development, Volpe researchers observed difficulties in calibrating their model, leaving them to question whether there existed flaws in their model, in the data, or in the procedure used to calibrate the model using the data. In this thesis, I use Bayesian methods for data analysis and parameter estimation to explore and, where possible, address these questions. 

First, I use Bayesian inference to measure the sufficiency of the size of the NDS data set. Second, I compare the procedure and results of the genetic algorithm-based calibration performed by the Volpe researchers with those of Bayesian calibration. Third, I explore the benefits of modeling car-following hierarchically. Finally, I apply what was learned in the first three phases using an established car-following model to the probabilistic modeling of WZDM. Validation is performed using information criteria as an estimate of predictive accuracy. A third model used for comparison with WZDM in the simulator, Wiedemann ’99, is also modeled probabilistically.

%% file: chap1.tex
\chapter{Introduction}	\label{chapter 1}

\section{Project Background and Motivation}

Traffic simulation software packages are widely used in transportation engineering to estimate the impacts of potential changes to a roadway network and forecast system performance under future scenarios. These packages are underpinned by math- and physics-based models, which are designed to describe behavior at an aggregate (macroscopic) level or the level of individual drivers (microscopic). Macroscopic models are used to evaluate aggregate measures, such as capacity, in large roadway networks, and model an entire traffic stream as a whole. Microscopic models are used to simulate the effects of local roadway elements on individual vehicles, and these individual effects can be aggregated to recover macroscopic effects if necessary, \cite{berthaume}. Within the microscopic realm, driver behavior is decomposed into sub-models that individually handle lane-changing (lateral movement), car-following (longitudinal movement), route choice, and so on. This work focuses specifically on car-following models (CFMs), which estimate the acceleration and deceleration behavior of individual vehicles with respect to their driving environment. Critical to the accurate performance of simulation models is the calibration process, during which unobserved model parameters have their values estimated using field measurements. Due to known variations in driver behavior that exist across regions of the roadway network (such as highway versus urban), across different roadway segment classes (e.g. signalized intersections versus roundabouts), and between different driving conditions (such as varying weather or construction), every CFM must be calibrated using data sampled locally from the area being studied to establish an accurate baseline before predicting the effects of changes to the local roadway configuration. 

One region of roadways where current CFMs have historically failed to accurately predict vehicle movements is work zones. The recognition of this fact has motivated the U.S. Department of Transportation's (USDOT's) Federal Highway Administration (FHWA) to fund research into the development of a new CFM that is effectively applicable to work zone regions as well as other regions of roadway networks. Part of that development involved the collection of field data for use in calibrating the model and demonstrating its efficacy.

Effective planning of work zone configurations that influence driver decision-making and behavior as transportation planners intend is crucial to the prevention of crashes, injuries, and fatalities in work zones, and to the mitigation of travel delays caused by work zones. FHWA partnered with researchers at USDOT's Volpe National Transportation Systems Center (Volpe) to address this problem. 

After constructing their proposed Work Zone Driver Model (WZDM) and conducting a naturalistic driving study (NDS) to collect time-series observations of the predictor and response variables relevant to that model, the Volpe researchers encountered challenges fitting the model to the observed data using the industry-standard automatic calibration method based on genetic algorithm (GA) search. Simulations drawn from the fit model produced unexpected vehicle behaviors, including crashes and rigid movements of platooned vehicles. These misbehaviors could have resulted from flaws in the model, flaws in the data or data collection process, flaws in the calibration procedure used to fit the model to the data, or some combination of the three. After discussing the problems with the researchers and reading the literature review that they performed to determine how and why they chose the exact procedure that they followed, I recognized that Bayesian methods for model parameter estimation, model validation, and model comparison were not considered at all. After the failure of the GA, the researchers resolved to hand-tuning parameter values based on their expert intuition and heuristics. Knowing that the researchers were working with a set of only approximately 200 car-following time-series instances and that Bayesian methods often outperform other machine learning methods when data set sizes are small, I recommended an investigation into the potential utility of following a Bayesian approach to CFM calibration and validation.

This project aims to determine whether approaches to calibration and validation based on Bayesian methods can resolve the aforementioned problems. The research agenda is divided into three phases: {\em Phase 1}: Formulation and Bayesian calibration of a probabilistic variant of a popular existing car-following model: the Intelligent Driver Model (IDM), \cite{treiber}, with subjective comparisons of procedures and results to GA-based calibration; {\em Phase 2}: Validation of calibration results and rigorous comparison of different probabilistic IDM formulations (pooled, hierarchical, and unpooled) with one another and with a GA-learned model; and {\em Phase 3}: Probabilistic formulation, calibration, and validation of the WZDM. 



% ...

\section{Research Questions}

The key question that this research is intended to address is whether the difficulties in calibration experienced by Volpe stem from problems with the data, the model, the calibration procedure, or perhaps all three. I hypothesize that the existing data set is sufficient for use in car-following model development given the use of a more appropriate calibration framework based on Bayesian methods. I have already shown in previous work that the challenges met when combining the NDS data with a GA-based method are not unique to the WZDM, but also present themselves when using the simpler and well-established IDM, \cite{abodo}. In that same work, I also qualitatively measured the size of the data set to be insufficiently large to dominate the influence of prior specification on Bayesian inference of model parameter values. Outstanding questions to be answered by this research effort include:

\begin{enumerate}
	\item In general, can a calibration procedure based on Bayesian inference produce measurably better results than one based on a genetic algorithm when available field data is low in quantity? 
	\item Given the use of Bayesian calibration, how do partially-pooled models that are structured hierarchically improve calibration results over fully pooled models when simulating the full population of car-following instances, if at all?
	\item How do the partially-pooled models improve on unpooled models when simulating per-framework or per-driver car-following instances, if at all? 
    \item Can a Bayesian calibration procedure provide a rigorous way to measure the sufficiency of the size of a data set?
    \item Can the Bayesian approach to model validation give a more convincing measure of a model's goodness of fit than the hypothesis tests traditionally used in car-following model calibration?
\end{enumerate} 

%% file: chap2.tex
\chapter{Related Work}		\label{chapter 2}

\section{Bayesian Methods Applied to Traffic Simulator Calibration}

Bayesian methods for traffic simulation model calibration, while dramatically less popular than methods based on optimization techniques, have received some attention in the last fifteen years. In earlier work on calibration for traffic simulators, Bayesian methods have been advocated because of their ability to capture uncertainty in estimated parameter values stemming from 1) inconsistency between the model and the natural phenomena it is meant to capture, 2) errors introduced by the calibration process itself (including the choices of measure of performance and measure of error), and 3) noise or errors in the data collection process, \cite{bayarri}, \cite{molina}. Bayesian methods for model comparison have also been applied, \cite{molenaar}.

In \cite{bayarri} and \cite{molina}, MCMC methods are used to analyze the error in human measurement of turn counts and roadway entry counts, and to estimate other parameters of the CORSIM simulator. In \cite{molina}, the influence of sampling from parameter distributions when generating simulation traces rather than fixing parameter values to distribution means or point estimates is additionally evaluated. \cite{zhong} modeled the Intelligent Driver Model using Gaussian random variables as the parameters to perform a probabilistic sensitivity analysis based on the Kullback-Liebler dissimilarity measure to limit the number of parameters requiring value estimation to those yielding the greatest performance improvement relative to default parameter values. \cite{rahman} advocated for the use of a Bayesian approach to car-following model calibration by directly comparing an MCMC-based calibration method with a deterministic optimization method, using one synthetic data set to show that the Bayesian method could recover known parameter values and one real-world data set as a case study. This work differs from that of  \cite{rahman} in that we leverage a hierarchical formulation of the models under study, in that our use of a probabilistic programming language allows us to formulate our models using distributions that are substantially more complex than the multivariate Gaussian used in \cite{rahman} can allow, including a Double Gamma output variable distribution. In \cite{molenaar}, the author augmented a physical traffic simulator so that background vehicles could adapt their simulated motions in near real-time to complement the driving behavior of the human operator using periodic updates to car-following model parameter estimates derived from observations of the operator. The objective was to remove bias from the operator's driving that might be induced by unrealistic surrounding vehicle motions. The simulator used was supported by the General Motors model and the Intelligent Driver Model, for each of which a multivariate Gaussian distribution was used to represent the joint distribution over the parameters, all of which are inherently continuous. The author further exploits the utility of model comparison in the Bayesian framework to measure the relative goodness of fit between the GM and IDM models given the data used to fit the models.


\section{Bayesian Analysis in Transportation Safety Assessment}

Much more common than Bayesian estimation of parameters for physical models in simulators is the estimation of statistical model parameters in traffic safety assessment, the ultimate purpose of the development of the Work Zone Driver Model. 

\cite{RAKHA2011739} used MCMC to fit a logistic Generalized Linear Model to observations of driver gap acceptance when making left turns. The authors judged the shape of the distributions over parameters estimated using the Bayesian method to represent a better model fit than those estimated using bootstrapping, an alternative statistical approach. 

With a purpose rather similar to ours, \cite{jimenez} used MCMC to estimate the parameters of a hierarchical Bayesian formulation of a pavement deterioration model using data collected from a specific region of interest. The authors further address the problem of missing data points in their time-series observations by applying Bayesian data imputation; using the available data to estimate posterior distributions for the missing data points. \cite{Inkoom} created a Bayesian version of a survival analysis model using the Weibull distribution to predict the time-to-failure of roadway pavement.

\cite{ankoor} used a 3-level hierarchical change-point model to determine if accident rates fell following the implementation of a change in how Los Angeles county ports manage freight truck traffic congestion. \cite{shaheed} developed a Bayesian hierarchical logit model to predict car crashes during winter based on season-specific weather attributes.

\section{Car-following Model Calibration Using Genetic Algorithms}

In their extensive report comparing the performance of different combinations of mathematical optimization techniques and goodness-of-fit measures, \cite{ciuffo} speculated that genetic algorithms may be the most commonly used class of optimization technique for calibrating traffic simulators.

\cite{Ranjitkar} used a genetic algorithm to evaluate the performance of six car-following models under different traffic conditions based on lead vehicle driving behavior. They observed that performance varied more as a function of driver behavior than of the choice of car-following model. 

\cite{hammit} develop a genetic algorithm as the basis for a general, transparent, and reproducible calibration procedure, using four car-following models from different classes to demonstrate generalizability: 1) collision avoidance, 2) psychophysical, 3) continuous response, and 4) low-order, piece-wise linear. Their results showed improved performance of the parameters that were calibrated to field data over default parameters. Two of those models, the Wiedemann '99 model, and Intelligent Driver Model, are used in this work as well.

Genetic algorithms are the foundation of GENOSIM, \cite{Ma}, a software tool for automatically calibrating the car-following, lane-changing, and gap-acceptance models that underlie microscopic traffic simulators. GENOSIM attempts to maximally generalize its applicability to a broad range of simulators and models by allowing its users to choose from four types of genetic algorithms that implement different parameter evolution strategies, acknowledging that genetic algorithms are "a problem-specific optimization technique" that require custom tuning. This work is intended to demonstrate Bayesian inference as a simpler approach that can produce better calibration results while requiring less configuration of the procedure.


%% file: chap3.tex
\chapter{Technical Background}		\label{chapter 3}

\section{Bayesian Programming}

A Bayesian program (BP) is a generic formalism that can be used to describe many classes of probabilistic models, including Hidden Markov Models, Bayesian Networks, and Markov Decision Processes, \cite{diard}. This formalism organizes a graphical model, which encodes prior knowledge about the inference problem, together with model variables and observed data into a structure like the following:
\newline
\begin{center}
\text{Program}
$\begin{cases}
\text{Description}
\begin{cases}
\text{Specification}
\begin{cases}
\text{Variables} \\
\text{Decomposition} \\
\text{Forms (parametric or program)}
\end{cases}\\
\text{Identification (using data)}
\end{cases}\\
\text{Questions.}
\end{cases}$
\newline
\end{center}

The program must define 1) a means of computing the joint probability over its model, variables, and data, and 2) a means of answering a specific inference question given that joint probability. For example, a Hidden Markov Model would have a sequence of states and observations as its variables, emission and transition matrices as its model, and various message-passing algorithms as its means of answering inference questions. The Baum-Welch algorithm, \cite{Baum}, would be its means of identification.

\section{Probabilistic Programming}

Probabilistic programming languages (PPLs) allow BPs to be implemented and have inference performed over their parameters using software. PPLs are like ordinary programming languages but they additionally allow variables' values to be randomly sampled from distributions, allowing the output of programs written in them to vary non-deterministically given the same input. They also allow variables' values to be conditioned on data (observations). In the BP formalism, the description of a probabilistic program is as follows: 
\newline
\begin{center}
\text{Program}
$\begin{cases}
\text{Description}
\begin{cases}
\text{Specification}
\begin{cases}
\text{Variables: $\theta$ \textit{(latent)} and \textbf{x} \textit{(obs)}} \\
\text{Decomposition: $P(\textbf{x}, \theta) \propto P(\theta) P(\textbf{x} | \theta)$} \\
\text{Form: \textit{Probabilistic Program}} \\
\end{cases}\\
\text{Identification: \textit{MCMC} or \textit{VI}}
\end{cases}\\
\text{Question: $P(\theta | \textbf{x}).$} \\
\end{cases}$
\newline
\end{center}

The decomposition states that the posterior joint probability of the model parameters and the observations is proportional to the product of the prior probability of the model parameters, $P(\theta)$, and the likelihood of the observations given the model parameters, $P(\textbf{x}|\theta)$. Because we treat each time step in each CF instance as independent, the likelihood can be further decomposed into products of the probabilities of each individual $x_i \in \textbf{x}$:
$$P(\textbf{x}|\theta) = \prod_{i=0}^{|\textbf{x}|} P(x_i|\theta).$$



The inference algorithms included with PPLs, such as Markov Chain Monte Carlo (MCMC) and variational inference (VI), can be run on arbitrary graphical models, as opposed to algorithms that run on a limited set of model classes for which they have been specially invented (e.g. Bayesian and Markov Networks). To construct a probabilistic variation of a deterministic mathematical model, such as a car-following model, one merely needs to implement the model as the likelihood function using the primitives of the PPL, with random variables (and their corresponding probability distributions) used to represent model parameters and the response variable, and ordinary mathematical operations used everywhere else. If the model can be implemented, then probabilistic parameter estimation can be performed in a plug-and-play fashion. The incredible flexibility of PPLs like Edward2 and PyMC3, \cite{pymc3}, allow models of varying compositions to be implemented, from the piece-wise linear Newell '02 model, \cite{newell}, to the highly non-linear Wiedemann '99 (W99) model, which includes conditional function evaluation. We chose the IDM model to demonstrate our calibration approach due to its moderate simplicity and interpretability, plus its widespread use in practice, \cite{hammit}.

\section{Bayesian Hierarchical Models}

When modeling a phenomenon for which observations naturally form a hierarchy or for which related sub-components are required, structuring the model hierarchically can improve estimation outcomes by introducing a form of regularization called partial pooling.

Consider the context of this research, wherein the goal is to model car-following. Some car-following models are designed under an assumption that their parameters would be fit using data from a single driver, \cite{treiber}. Others are only theoretically valid when fit using a single instance, \cite{newell}. In some cases, when available data are not plentiful, these assumptions are violated and a single global model is fit using the entire data set in order to best approximate the true distribution of behavior over all drivers or instances given a sample of drivers or instances, \cite{ear}. Rather than having to forego the desired per-driver/instance models in exchange for a more informed estimate (using a fully pooled model), or forego the benefit of using all data points to inform estimation in exchange for individual models (using fully unpooled models), one can achieve both by using a hierarchical model that implements partial pooling.

In the Bayesian framework, where latent model parameters are random variables with distributions, partial pooling is implemented by having the priors of the individual sub-components at a base level (level-1) be drawn from a shared set of distributions (at level-2). This introduces a conditional dependency between the subgroups that leads each sub-model's estimation to be informed by all data points while remaining distinct from other models' estimations. The regularizing effect applies in two directions. 

For level-1 parameters, the means are pulled toward the mean of the corresponding level-2 parameters in an effect known as shrinkage, or Bayesian smoothing, \cite{gelfand}. This is especially useful when it is known that the level-1 distributions should not be too dissimilar. For example, drivers have their personal driving preferences but are all still human drivers. Further, the degree to which shrinkage occurs for each component tends to vary proportionally to the amount of data supporting that component. 

For level-2 parameters, the means are also adjusted by comparison with a fully pooled model. The use of hierarchy decouples the variance that can be attributed to each sub-population in level-1 from that of the full population in a fully pooled scenario. The effect is an estimate that is closer to the true population than the sample implies. As with other methods of regularization, the predictive accuracy of higher-level hierarchical parameters is expected to be worse than fully pooled parameters, which is a desirable outcome when assuming that a data set is not properly representative of the phenomenon being modeled.

Hierarchical models can have an arbitrarily high number of levels, and can even support partial pooling across multiple dimensions at the same level for data that are compositional (i.e. data points that belong to more than one category at the same level of the hierarchy), \cite{ mcelreath}. 

\section{Markov chain Monte Carlo and Metropolis Hastings}

Markov chain Monte Carlo (MCMC), \cite{barnes}, is a family of simulation algorithms that allow the integrals common to the probability density functions of complex and high-dimensional distributions, which do not have closed-form analytical solutions, to be approximated. This ability enables Bayesian inference to be performed for models of natural phenomena like car-following, and for models implemented as probabilistic programs to be composed of arbitrary combinations of prior and likelihood distributions without having to consider conjugacy of the prior with respect to the likelihood.

Metropolis Hastings, \cite{barnes}, a popular MCMC algorithm that samples a sequence of states in a state space, called a Markov chain, from a candidate probability density function. Transitions from one state to the next are determined using an acceptance test applied to proposals for the next state such that, after a number of iterations, the region of the state space where the chain ends corresponds to the desired density function. In this work, the random walk variant of Metropolis Hastings is used, wherein proposals are generated by sampling from a Gaussian distribution with the current state as the mean. The algorithm can be considered to have converged when a sufficiently long sequence of states stabilizes within a fixed region.  

\section{Evolutionary Algorithms and Differential Evolution}

Evolutionary algorithms are a class of global optimization algorithms that derive their search strategies from concepts in evolutionary biology, such as mutation, crossover, and selection within a population, and include Genetic Algorithms, \cite{Mitchell}. They apply to problems where the search space is non-convex or non-differentiable, and are thus an attractive option for calibration of car-following models, which often compose discrete sub-models. 

Evolutionary algorithms are typically initialized with a random selection of population members, where each member is a parameter vector. Over the course of the search, the values of the vectors are perturbed according to mutation and crossover rules, and new members of the population are selected to replace existing ones based on some fitness function to create the next in a sequence of generations. This use of populations and random perturbations enable evolutionary algorithms to avoid being trapped in local minima of the search space, \cite{Ranjitkar}.

Differential Evolution (DE), \cite{Kara}, is one instance of this class of algorithms that is used in this work. With DE, to generate the next generation of the population, candidate vectors are constructed by mutating two existing population vectors, performing crossover between the result and a third target vector, then conditionally selecting the candidate for inclusion if its fitness is higher than the original vector. The algorithm terminates when all members of the population converge to the same value within some error.

\section{Bayesian Model Validation using Information Criterion}

The standard best practice for validating statistical and machine learning models is to perform either \textit{K}-fold or leave-one-out (LOO) cross-validation (CV), especially when a data set is too small to expect a held-out test set to be equally representative of the data distribution as the training set. Cross-validation estimates a model's predictive accuracy on out-of-sample data by partitioning the total data set into \textit{K} subsets (with \textit{K} = 1 corresponding to LOO-CV),  performing \textit{K} distinct model fits with a disjoint subset held out each time, using that held-out data subset to measure the \textit{K}th model's performance, and finally averaging over the \textit{K} models' performance metrics. The Bayesian framework offers validation metrics called \textit{information criterion} that asymptotically approximate LOO-CV using the entire data set, which is valuable when the data set size is small. Two criterion of particular interest are 1) the Watanabe-Akaike information criterion (WAIC), \cite{Watanabe}, which operates on samples from the posterior log-likelihood distribution one observation at a time, and then sums over all observations in the data set to yield a single measure, and 2) Pareto Smoothed Importance Sampling-based LOO (PSIS-LOO) CV, which improves on WAIC "in the finite case with weak priors or influential observations", \cite{Vehtari2017}, which is precisely our scenario. 

%% file: chap4.tex
\chapter{Car-following Models}	\label{chapter_4}

In this chapter, the three car-following models used in this research are introduced. The primary objective of the project was to demonstrate the utility of modeling and fitting the Work Zone Driver Model using probabilistic programming. That effort involved comparing the performance of the WZDM with that of the model it is intended to improve upon in work zone environments, the Wiedemann '99 model. To gauge the potential of the research direction, experiments were first performed using the Intelligent Driver Model because its relative simplicity made it easy to understand, implement and analyze.

\section{The Intelligent Driver Model}

The Intelligent Driver Model, \cite{treiber}, predicts a following vehicle's next acceleration, $a_{t+1}$, given that vehicle's absolute velocity, $v$, and its velocity and distance relative to a leading vehicle, $\Delta v$ and $s$, at the current time:
\newline
\begin{equation}
    a_{t+1} = a\bigg[1 - \bigg(\frac{v}{v_0}\bigg)^{\delta} - \bigg(\frac{s^*(v,\Delta v)}{s}\bigg)^2\bigg],
\end{equation}
\newline
with:
\newline
\begin{equation}
    s^*(v,\Delta v) = s_0 + s_1 \sqrt{\frac{v}{v_0}} + \textit{T}v + \frac{v \Delta v}{2 \sqrt{a b}}.
\end{equation}
\newline
The estimable parameters are $v$, the desired velocity; $T$, the "safe" time headway, which is the time it would take the following vehicle to close the gap between itself and the leading vehicle; $a$, the maximum acceleration; $b$, the desired or "comfortable" deceleration, with higher values corresponding to more aggressive and late braking; $s_0$ and $s_1$, jam distance terms that correspond to stop-and-go traffic when values are low; and $\delta$, which represents the rate at which a driver will decrease acceleration as the desired velocity is approached.  This decrease in acceleration can range anywhere from exponential to linear or sub-linear.

IDM is a member of the class of collision-free car-following models. In ordinary situations, the vehicle will decelerate according to $b$, but in emergencies, deceleration can occur at an exponential rate, \cite{ferreira}. Other widely used car-following models include the Gipps model, \cite{gipps}, which is used in the AIMSUN simulator and is also a collision-free model, and Wiedemann '99, a psychophysical model that is the basis of VISSIM, \cite{barcelo}. IDM is itself included as an option in the deep reinforcement learning framework for traffic simulation, Flow, \cite{flow}.

\section{The Wiedemann 99' Model}

The Wiedemann 99' Model, \cite{wiedemann}, is a psychophysical model that predicts a following vehicle's next acceleration, $a_{t+1}$, given that vehicle's velocity and distance relative to a leading vehicle, and several other parameters that characterize the following driver's preferences. The theory asserts that drivers alternate between states of conscious reaction, unconscious reaction, and no reaction to lead vehicles. The model parameters and equations utilized depend on which one of four reaction regimes the driver is operating in. In the \textit{freeflow} regime, the lead vehicle is sufficiently far away for the following driver to accelerate to and maintain their preferred speed. This is a no-reaction scenario. In the \textit{approaching} regime, the follower consciously reacts by decelerating at a moderate rate until a preferred following distance is achieved. The \textit{danger} regime corresponds to an emergency braking situation in which the driver consciously decelerates at a high rate to avoid a collision. The \textit{following} regime is the hallmark of this model. In this regime, the subject driver subconsciously alternates between mild acceleration and deceleration to maintain a preferred distance. W99 model equations are listed below.

W99 model equations:

\begin{equation}
a_{t+1} = 
\begin{cases}
\begin{array}{ l l }
a_{f} = a_{following} & \text{if} \quad R(t) = R_{f} = R_{following} \\
a_{a} = a_{approaching} & \text{if} \quad R(t) = R_{a} = R_{approaching} \\
a_{d} = a_{danger} & \text{if} \quad R(t) = R_{d} = R_{danger} \\
a_{ff} = a_{freeflow} & \text{if} \quad R(t) = R_{ff} = R_{freeflow} \\
\end{array}
\end{cases}
\end{equation}

\begin{equation}
R(t) = 
\begin{cases}
\begin{array}{ l l }
R_{f} & \text{if} \quad SDV_o \geq {\Delta}v \wedge SDV_c \leq {\Delta}v \wedge SDX_o \geq {\Delta}x \wedge SDX_c < {\Delta}x \\
R_{a} & \text{if} \quad SDV_c > {\Delta}v \wedge DSXV > {\Delta}x \wedge SDX_c < {\Delta}x \\
R_{d} & \text{if} \quad SDX_c \geq {\Delta}x \wedge SDV_o \geq {\Delta}v \\
R_{ff} & \text{otherwise} \\
\end{array}
\end{cases}
\end{equation}

W99 framework equations:

\begin{equation}
SDV_o = 
\begin{cases}
\begin{array}{ l l }
CC_5 + CC_6{\Delta}x^2 & \text{if} \quad v_F > CC_5 \\
CC_6{\Delta}x^2 & \text{otherwise} \\
\end{array}
\end{cases}
\end{equation}

\begin{equation}
SDV_c = 
\begin{cases}
\begin{array}{ l l }
CC_4 - CC_6{\Delta}x^2 & \text{if} \quad v_L > 0 \\
0 & \text{otherwise} \\
\end{array}
\end{cases}
\end{equation}

\begin{equation}
SDXV = SDX_o + CC_3({\Delta}v - CC_4)
\end{equation}

\begin{equation}
SDX_o = CC_2 + SDX_c
\end{equation}

\begin{equation}
SDX_c = 
\begin{cases}
\begin{array}{ l l }
CC_0 + CC_1v_F & \text{if} \quad v_L \geq 0 \wedge ({\Delta}v \geq 0 \vee a_L < -1) \\
CC_0 + CC_1(v_L - 0.5{\Delta}v) & \text{if} \quad v_L \geq 0 \wedge {\Delta}v < 0 \wedge a_L \geq -1 \\
CC_0 & \text{otherwise} \\
\end{array}
\end{cases}
\end{equation}

W99 regime equations:

\begin{equation}
a_{f} = 
\begin{cases}
\begin{array}{ l l }
min(max(a_F, CC_7), \frac{v_0 - v_F}{T}) & \text{if} \quad a_F > 0 \\
min(a_F, -CC_7) & \text{otherwise} \\
\end{array}
\end{cases}
\end{equation}

\begin{equation}
a_{a} = max(\frac{0.5{\Delta}v^2}{SDX_c - {\Delta}x - 0.1}, -10)
\end{equation}

\begin{equation}
a_{d} = 
\begin{cases}
\begin{array}{ l l }
0 & \text{if} \quad a_{d}* = 0 \\
-CC_7 & \text{if} \quad a_{d}* > -CC_7 \\
max(a_{d}*, 0.5\sqrt{v_F} - 10) & \text{if} \quad a_{d}* \leq -CC_7 \\
\end{array}
\end{cases}
\end{equation}

\begin{equation}
a_{d}* = 
\begin{cases}
\begin{array}{ l l }
0 & \text{if} \quad v_F \leq 0 \\
min(a_L + \frac{{\Delta}v^2}{CC_0 - {\Delta}x}, a_F) & \text{if} \quad v_F > 0 \wedge {\Delta}v < 0 \wedge {\Delta}x > CC_0 \\
min(a_L + 0.5({\Delta}v - SDV_o), a_F) & \text{if} \quad v_F > 0 \wedge {\Delta}v < 0 \wedge {\Delta}x \leq CC_0 \\
\end{array}
\end{cases}
\end{equation}

\begin{equation}
a_{ff} = 
\begin{cases}
\begin{array}{ l l }
0 & \text{if} \quad a_{ff}* = 0 \\
min(a_{ff}*, \frac{v_0 - v_F}{T})& \text{otherwise} \\
\end{array}
\end{cases}
\end{equation}

\begin{equation}
a_{ff}* = 
\begin{cases}
\begin{array}{ l l }
0 & \text{if} \quad SDX_c \geq {\Delta}x \\
CC_7 & \text{if} \quad R(t-1) = R_{ff} \wedge SDX_c < {\Delta}x \\
min(\frac{{\Delta}v^2}{SDX_o - {\Delta}x}, a_{u}) & \text{if} \quad R(t-1) \neq R_{ff} \wedge SDX_c < {\Delta}x \wedge SDX_o > {\Delta}x \\
a_{u} & \text{if} \quad R(t-1) \neq R_{ff} \wedge SDX_c < {\Delta}x \wedge SDX_o \leq {\Delta}x \\
\end{array}
\end{cases}
\end{equation}

\begin{equation}
a_{u} = CC_8 + min(v_F, v_{u})CC_9
\end{equation}

\begin{equation}
v_{u} = 22.22
\end{equation}

\begin{table}
\begin{center}
\caption {W99 Observed Variables} 
\label{tab:observed_variables}
\centering
\begin{tabular}{|l|l|l|}
\hline
\textbf{Name} & \textbf{Units} & \textbf{Description} \\
\hline
$v_F$ & m/s & Velocity of the subject vehicle \\
\hline
$a_F$ & $m/s^2$ & Acceleration of the subject vehicle \\
\hline
$v_L$ & m/s & Velocity of the vehicle in front of the subject \\
\hline
$a_L$ & $m/s^2$ & Acceleration of the vehicle in front of the subject \\ 
\hline
$\Delta v$ & m/s & Difference between lead and following vehicle velocities \\
\hline
$\Delta x$ & m & Distance between the lead and following vehicles \\
\hline
\end{tabular}
\end{center}
\end{table}

\begin{table}
\begin{center}
\caption {W99 Estimable Parameters} 
\label{tab:w99_model_parameters}
\centering
\begin{tabular}{|l|l|l|}
\hline
\textbf{Name} & \textbf{Units} & \textbf{Description} \\
\hline
$CC_0$ & m & Standstill distance \\
\hline
$CC_1$ & s & Spacing time \\
\hline
$CC_2$ & m & Following variation \\
\hline
$CC_3$ & s & Following entrance threshold \\ 
\hline
$CC_4$ & m/s & Negative following threshold \\
\hline
$CC_5$ & m/s & Positive following threshold \\
\hline
$CC_6$ & rad/s & Oscillation speed dependency \\
\hline
$CC_7$ & $m/s^2$ & Oscillation acceleration \\
\hline
$CC_8$ & $m/s^2$ & Standstill acceleration \\ 
\hline
$CC_9$ & $m/s^2$ & Acceleration at 80km/h \\
\hline
$v_0$ & m/s & Desired velocity \\
\hline
\end{tabular}
\end{center}
\end{table}

\begin{table}
\begin{center}
\caption {W99 Model Variables and Constants} 
\label{tab:w99_model_variables_and_constants}
\centering
\begin{tabular}{|l|l|l|}
\hline
\textbf{Name} & \textbf{Units} & \textbf{Description} \\
\hline
$a_u$ & $m/s^2$ & Acceleration upper bound \\
\hline
$v_{u}$ & m/s & Velocity upper bound \\
\hline
$T$ & s & Timestep duration \\
\hline
\end{tabular}
\end{center}
\end{table}

\section{The FHWA Work Zone Driver Model}

\begin{figure}[!t]
\centering
\includegraphics[width=5.8in]{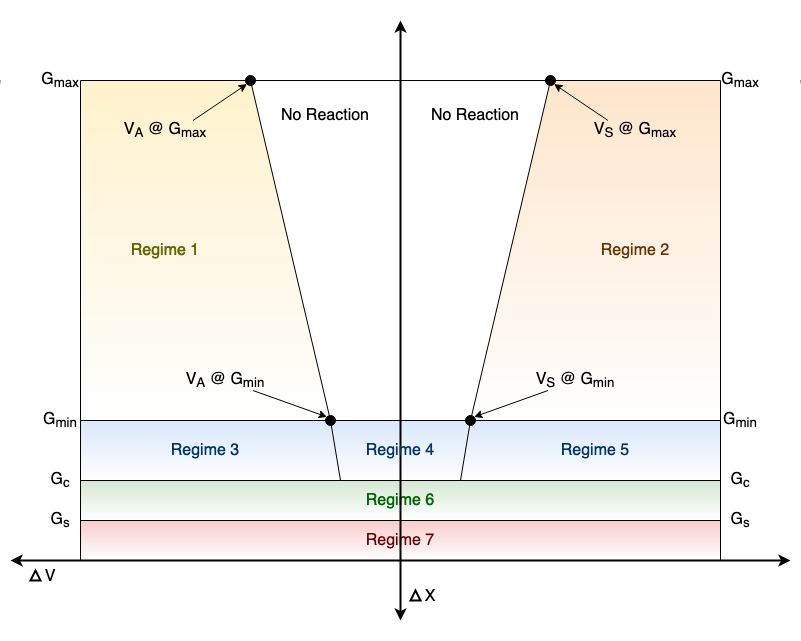}
\caption{The forces and relations between them that govern the value of acceleration at the next time step are determined by the (${\Delta}v$, ${\Delta}x$) location on the psychophysical plane at the current time step.}
\label{fig:wzdm_psychophysical_plane}
\end{figure}

The FHWA Work Zone Driver Model (WZDM), \cite{wzdm}, is a psychophysical model that predicts a following vehicle's next acceleration, $a_{t+1}$, given that vehicle's velocity and distance relative to a leading vehicle, and several other parameters that characterize the following driver's preferences, similarly to the W99 model. The relations between these variables are modeled using force equations inspired by Modified Field Theory from the field of psychology, \cite{berthaume}. WZDM has seven driver response regimes, each of which combines different force equations linearly. The regime at time $t$, $R(t)$, is determined by the location of the point (${\Delta}v$, ${\Delta}x$) on the psychophysical plane depicted in Figure \ref{fig:wzdm_psychophysical_plane}.

\begin{table}[ht]
\begin{center}
\caption {WZDM Framework IDs} 
\label{tab:wzdm_framework_ids}
\centering
\begin{tabular}{|l|l|l|}
\hline
 & \textbf{Congested} & \textbf{Uncongested} \\
\hline
\textbf{Highway} & 1 & 2\\
\hline
\textbf{Freeway} & 3 & 4\\
\hline
\textbf{Advanced Warning} & 5 & 6 \\
\hline
\textbf{Taper Zone} & 7 & 8 \\
\hline
\textbf{Work Zone with Lane Closure} & 9 & 10 \\
\hline
\textbf{Work Zone without Lane Closure} & 11 & 12 \\
\hline
\end{tabular}
\end{center}
\end{table}

The strategy that the WZDM employs to improve model performance in work zones is to allow parameter values to vary depending on which one of 12 scenarios called frameworks the driver is in. The frameworks are composed of two dimensions of variation: facility type and congestion level, and each has its own set of dedicated framework parameters and driving regime parameters. Table \ref{tab:wzdm_framework_ids} enumerates each framework. Freeways are a sub-type of highways that have controlled entrances and exits (e.g. on- and off-ramps), typically higher speed limits, no intersections, and barriers between traffic lanes of opposing direction. Advanced warning facilities contain construction signs that indicate the distance to an approaching work zone. Taper zones provide gradual transitions into and out of work zone regions with lane closures. A third dimension of variation exists in the original model named \textit{lead vehicle type} that includes the values passenger and heavy vehicle. Since only passenger types are represented in the data set used for calibration, this third dimension is not considered in this work. Framework and regime parameters are listed in Table \ref{tab:wzdm_framework_parameters} and Table \ref{tab:wzdm_regime_parameters}. Observed variables are presented in Table \ref{tab:wzdm_observed_variables}.

WZDM model equations:

\begin{equation}
a_{t+1} = 
\begin{cases}
\begin{array}{ l l }
a_{case1} & \text{if} \quad R(t) = R_{1} \vee R_{2}\\
a_{case2} & \text{if} \quad R(t) = R_{3} \vee R_{4} \vee R_{5} \\
a_{case3} & \text{if} \quad R(t) = R_{6} \\
a_{case4} & \text{if} \quad R(t) = R_{7} \\
a_t & \text{otherwise}
\end{array}
\end{cases}
\end{equation}

Regime equations:
\begin{equation}
a_{case1} = F_{lead-vel} + F_{des-vel} + F_{des-prox} + F_{\frac{{\Delta}x}{{\Delta}v}} + F_{BL} + F_{gap}
\end{equation}
\begin{equation}
a_{case2} = 2(F_{lead-vel} + F_{des-prox}) + \frac{F_{des-vel}}{1.5} + F_{\frac{{\Delta}x}{{\Delta}v}} + F_{BL} + F_{gap}
\end{equation}
\begin{equation}
a_{case3} = 
\begin{cases}
\begin{array}{ l l }
A_{des-max-decel} & \text{if} \quad V_L \geq 5 \\
2(F_{lead-vel} + F_{prox}) & \text{otherwise} \\
\end{array}
\end{cases}
\end{equation}
\begin{equation}
a_{case4} = A_{des-emergency-decel}
\end{equation}

Force equations:
\begin{equation}
F_{lead-vel}(t) = C_v\frac{V_{L,t-PRT{\Delta}v} - V_{F,t-PRT{\Delta}v}}{|{\Delta}x_{t-PRT{\Delta}x} - G_c|}
\end{equation}

\begin{equation}
F_{des-vel}(t) = C_{des}(V_{des} - V_{F,t-PRT{\Delta}v})
\end{equation}

\begin{equation}
F_{des-prox}(t) = \frac{-C_{prox}N}{|{\Delta}x_{t-PRT{\Delta}x} - G_s|}
\end{equation}

\begin{equation}
F_{gap}(t) = 
\begin{cases}
\begin{array}{ l l }
-C_{gap}\frac{{\Delta}v}{\Delta x} & \text{if} \quad \frac{{\Delta}x}{V_F} < {\Delta}x \\
0 & \text{otherwise} \\
\end{array}
\end{cases}
\end{equation}

\begin{equation}
F_{BL}(t) = 
\begin{cases}
\begin{array}{ l l }
-C_{BL}\frac{A_L}{\sqrt{{\Delta}x}} & \text{if} \quad A_L < 0 \\
0 & \text{otherwise} \\
\end{array}
\end{cases}
\end{equation}

\begin{equation}
F_{\frac{{\Delta}x}{{\Delta}v}}(t) = 
\begin{cases}
\begin{array}{ l l }
-2 & \text{if} \quad 0 < \frac{{\Delta}x}{{\Delta}v} < TTC_{extreme}\\
0 & \text{otherwise} \\
\end{array}
\end{cases}
\end{equation}

Constants:
\begin{equation}
leader\_velocity\_threshold = 2.2352
\end{equation}
\begin{equation}
extreme\_time\_to\_collision = 6.
\end{equation}
\begin{equation}
extreme\_approach\_acceleration = -2.
\end{equation}

\begin{table}
\begin{center}
\caption {WZDM Observed Variables} 
\label{tab:wzdm_observed_variables}
\centering
\begin{tabular}{|l|l|l|}
\hline
\textbf{Name} & \textbf{Units} & \textbf{Description} \\
\hline
$v_F$ & m/s & Velocity of the subject vehicle \\
\hline
$a_F$ & $m/s^2$ & Acceleration of the subject vehicle \\
\hline
$v_L$ & m/s & Velocity of the vehicle in front of the subject \\
\hline
$a_L$ & $m/s^2$ & Acceleration of the vehicle in front of the subject \\ 
\hline
$\Delta v$ & m/s & Difference between lead and following vehicle velocities \\
\hline
$\Delta x$ & m & Distance between the lead and following vehicles \\
\hline
\end{tabular}
\end{center}
\end{table}

\begin{table}
\begin{center}
\caption {WZDM Estimable Framework Parameters} 
\label{tab:wzdm_framework_parameters}
\centering
\begin{tabular}{|l|l|l|}
\hline
\textbf{Name} & \textbf{Units} & \textbf{Description} \\
\hline
$N$ & m & Force field shaping variable for proximity to passenger car \\
\hline
$C_v$ & m/s & Adjusts the influence of relative speed to the lead vehicle  \\
\hline
$C_{des}$ & $\frac{1}{s}$ & Adjusts the influence of desired speed \\
\hline
$C_{prox}$ & $m/s^2$ & Adjusts the influence of proximity to passenger car \\
\hline
$PRT_{{\Delta}v}$ & s & Perception-reaction time for relative velocity \\
\hline
$PRT_{{\Delta}x}$ & s & Perception-reaction time for relative distance \\
\hline
$PRT_v$ & s & Perception-reaction time for following vehicle’s velocity \\
\hline
$C_{gap}$ & m/s & Calibration parameter to adjust the influence of desired gap \\
\hline
$T_{safe}$ & s & Desired safe following time gap \\
\hline
$C_{BL}$ & $\sqrt{\text{m}}$ & Adjusts the influence of the brake lights of the lead vehicle \\
\hline
$V_{des}$ & m/s & Desired velocity \\
\hline
$A_{max}$ & $m/s^2$ & Desired maximum acceleration \\
\hline
$D_{max}$ & $m/s^2$ & Desired maximum deceleration \\
\hline
$D_{emr}$ & $m/s^2$ & Desired emergency deceleration \\
\hline
\end{tabular}
\end{center}
\end{table}

\begin{table}
\begin{center}
\caption{WZDM Estimable Regime Parameters}
\label{tab:wzdm_regime_parameters}
\centering
\begin{tabular}{|l|l|l|}
\hline
\textbf{Name} & \textbf{Units} & \textbf{Description} \\
\hline
$G_{min}$ & m & The minimum relative distance during car-following \\
\hline
$G_{max}$ & m & The maximum relative distance during car-following \\
\hline
$G_c$ & m & The absolute minimum distance gap \\
\hline
$G_s$ & m & Distance gap at stopped position \\ 
\hline
$V_{s,G_{max}}$ & m/s & Separating velocity at $G_{max}$ \\
\hline
$V_{s,G_{min}}$ & m/s & Separating velocity at $G_{min}$ \\
\hline
$V_{a,G_{max}}$ & m/s & Approaching velocity at $G_{max}$  \\
\hline
$V_{a,G_{min}}$ & m/s & Approaching velocity at $G_{min}$ \\
\hline
\end{tabular}
\end{center}
\end{table}

%% file: chap5.tex
\chapter{Data Description and Analysis}	\label{chapter 5}

In this chapter, the data supporting this research are described and analyzed to develop a general understanding and to inform modeling decisions. Based on the analysis of the sole dependent variable, follower acceleration, it was determined that the best distribution to model follower acceleration is the Double Gamma (also known as Reflected Gamma or Two-sided Gamma). The Double Gamma generalizes the Laplace distribution by introducing a shape parameter. The Gaussian distribution was chosen to model all latent variables. The descriptive statistics also motivate the exploration of hierarchical Bayesian models performed in this research by showing that fragmentation of the data into subsets per framework leads to a highly non-uniform distribution of data across frameworks.

\section{Data Description} \label{sec:data_description}

Those latent variables in a car-following model that are meant to capture driver preferences, and to make the model robust to a variety of preferences, must be estimated using real-world data collected in the geographic region and along the roadway segment types where analysis is to be performed. The road segments of interest for the WZDM are highway, freeway, advanced warning zone, taper zone, work zone with lane closure, and work zone without lane closure. To collect field measurements that correspond to the model's observed variables, which are presented in Table \ref{tab:wzdm_observed_variables} in the previous chapter, a research vehicle was outfitted with radar, GPS, video camera, and other sensors, and driven around the Greater Boston metropolitan area in Massachusetts, USA. The data were collected at a rate of 10 hertz by 53 participant drivers using one vehicle. 

The WZDM data are naturally organized into a hierarchical structure. Individual data points measured at each time step constitute car-following instance time series. By construction, each instance belongs in its entirety to one framework. Each instance also belongs to one driver. Each set of framework instances intersects with one or more sets of driver instances, and both are sub-groups of the total population of all instances.

\section{Data Analysis}
\label{sec:data_analysis}

The data analysis presented in this section motivated the decision to treat the data points at each time step as being independent and identically distributed (i.i.d.), to model the dependent variable using a Double Gamma random variable, and to model continuous parameters using Gaussian random variables. Discrete parameters are discussed in Chapter \ref{chapter 6}.

\begin{figure}
\begin{center}
\includegraphics[width=6in]{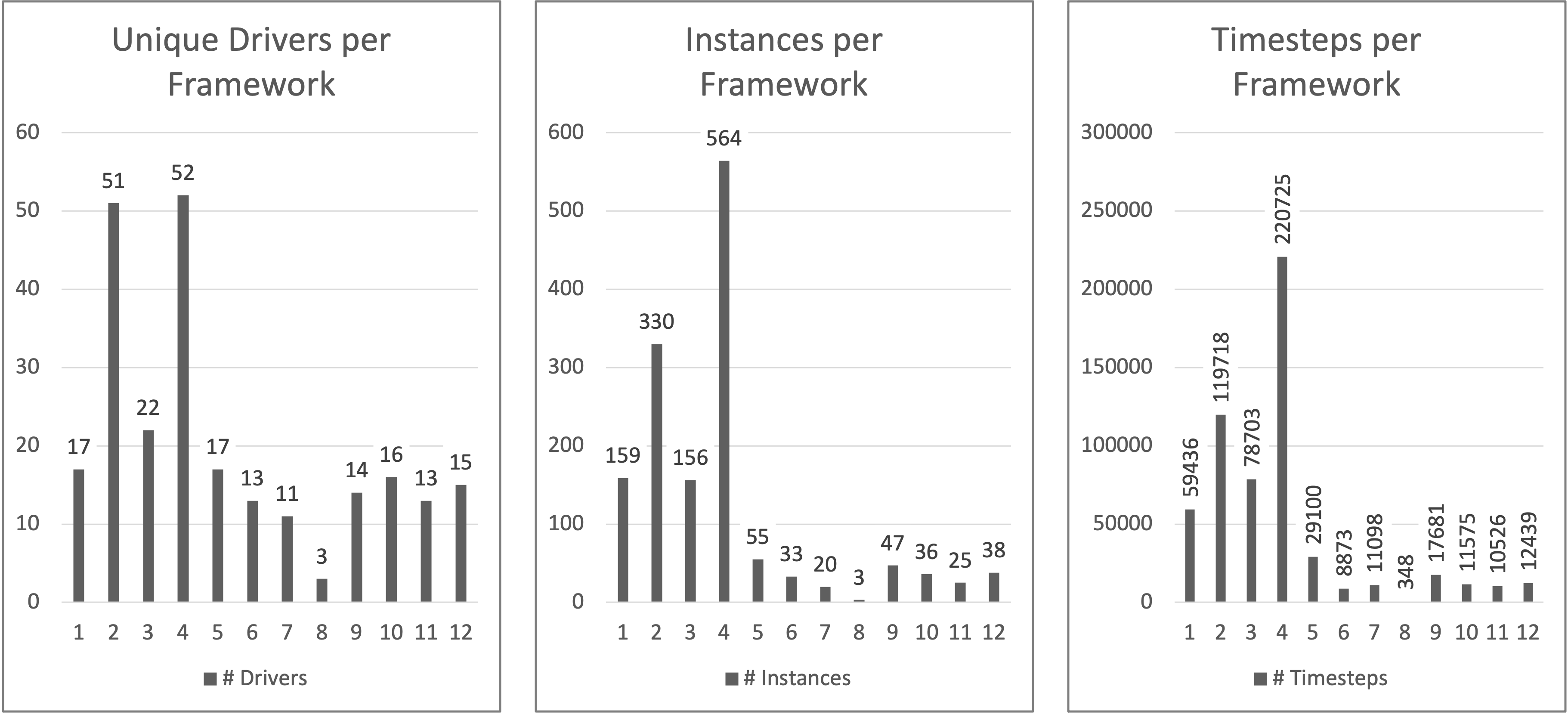}
\caption{Data frequency distributions over the set of car-following frameworks.}
\label{fig:per_framework_histograms}
\end{center}  
\end{figure}

\subsection{Analysis of Car-following Instance Frequency}
\label{subsec:analysis_of_car_following_instance_frequency}

\begin{figure} 
\begin{center}  
\includegraphics[width=6in]{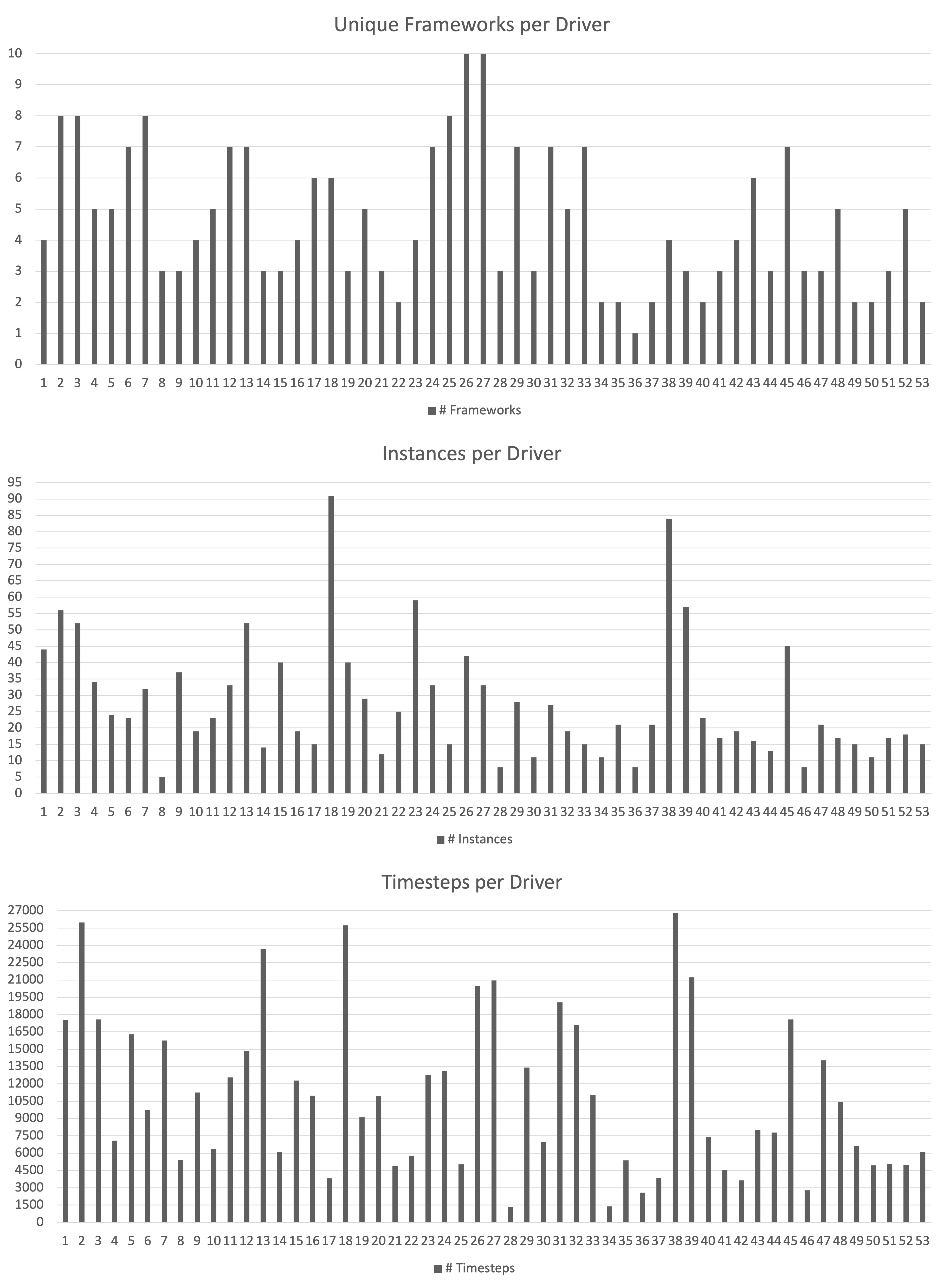}
\caption{Data frequency distributions over the set of drivers.}
\label{fig:per_driver_histograms}
\end{center}  
\end{figure}

The frequency distributions of driver count, CF instance count, and total timestep count per WZDM framework are illustrated in Figure \ref{fig:per_framework_histograms}. Each distribution is highly non-uniform, revealing that most frameworks have observations contributed to them by a small subset of drivers. In an extreme case, data belonging to Framework 8 is contributed by only three of the 53 drivers, and each of those three drivers contributes just one CF instance having an average length of only 116 timesteps. The longest CF instances in the data set have 20-to-30 times the number of data points. By comparison, Framework 4 has representation from 52 of 53 drivers, and 564 CF instances each having an average of 391 timesteps. No frameworks have data contributed by all drivers. This imbalance in the distribution of data across frameworks strongly motivates the use of hierarchical structure in the probabilistic CF models, which is expected to discourage over-fitting that might be detrimental to the models' predictive performance by allowing the data from different subgroups to inform one another's parameter estimations. 

\begin{figure}
\begin{center}  
\includegraphics[width=6in]{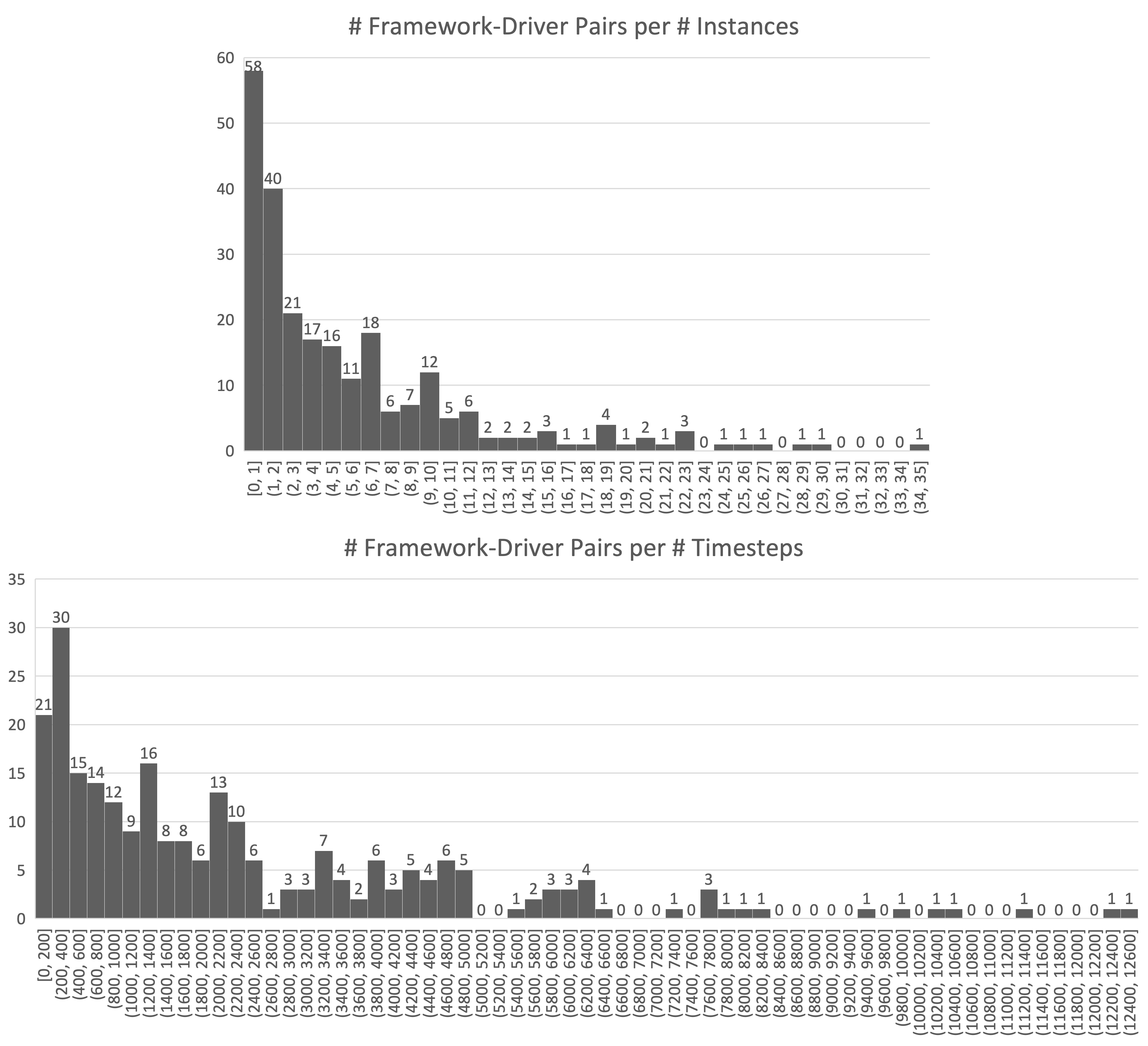}
\caption{Data frequency distributions over the number of instances and the number of timesteps for each intersection of framework and driver.}  
\label{fig:per_framework_driver_pair_histograms}
\end{center}  
\end{figure}

Figure \ref{fig:per_driver_histograms} shows the distributions of framework count, CF instance count, and total timestep count per driver. Again, the data are not balanced. There are 244 unique combinations of driver and framework in the data set. Figure \ref{fig:per_framework_driver_pair_histograms} displays the instance and timestep histograms for the combinations.

\subsection{Analysis of Observed Variable Distribution Shapes}
\label{subsec:analysis_of_observed_variable_distribution_shapes}

\subsubsection{Input Variable Distributions}
\label{subsubsec:input_variable_distributions}

\begin{figure} 
\begin{center}  
\begin{tabular}{c} 
\includegraphics[width=6in]{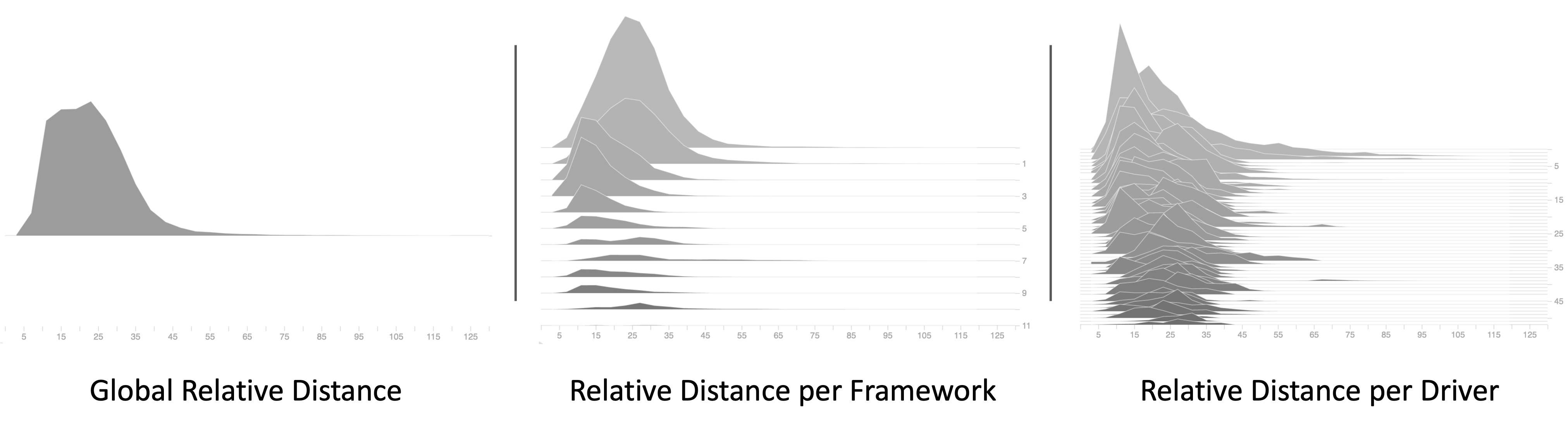}\\
(a)
\end{tabular}  
\begin{tabular}{c} 
\includegraphics[width=6in]{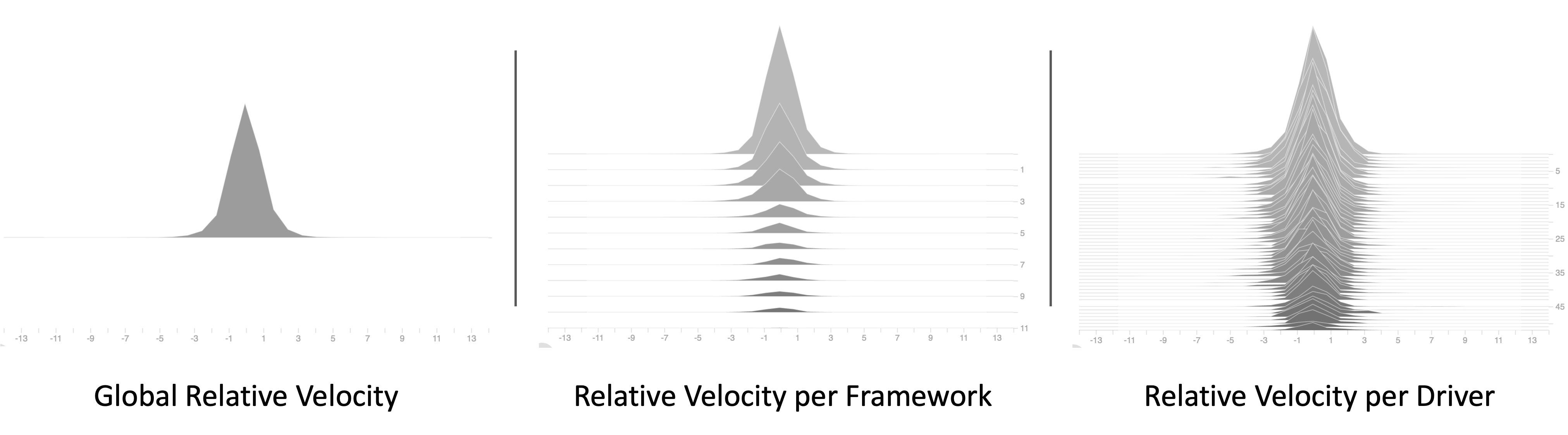}\\
(b)
\end{tabular}  
\begin{tabular}{c} 
\includegraphics[width=6in]{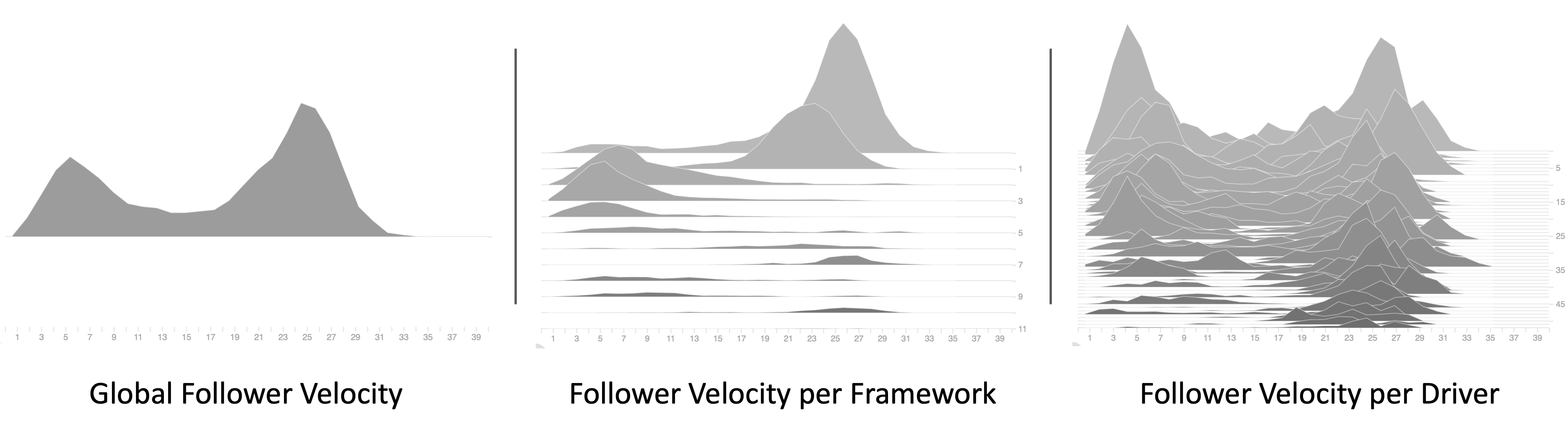}\\
(c)
\end{tabular}  
\caption{Histograms of the input variables relative distance (a), relative velocity (c), and follower velocity (c) for the global population, sub-populations per framework, and sub-populations per driver.}
\label{fig:relative_distance_histograms_and_relative_velocity_histograms}
\end{center}  
\end{figure}

Relative distance and relative velocity are the most fundamental input variables of psychophysical car-following models like the WZDM. Follower velocity is a variable that encodes more information about a driver's style and preferences while in motion compared with acceleration, which is almost always zero or close to zero. By analyzing the shapes of the distributions of these three variables, the selection of an appropriate distribution for modeling the latent variables can be informed. 

The histograms in Figure \ref{fig:relative_distance_histograms_and_relative_velocity_histograms} (a) of relative distance over the total population, over each framework, and over each driver show single modes (with one exception) that have sharp or rounded peaks and concave-in sides. They are also mostly left-skewed. Under ordinary conditions, the relative distance is expected to always be positive, meaning the support of the distribution should not include negative numbers. These properties could correspond to a Gamma distribution or, less exactly, a Gaussian modified by a bijection that constrains all values to be positive. 

The relative velocity distributions in Figure  \ref{fig:relative_distance_histograms_and_relative_velocity_histograms} (b) have sharp peaks and, concave-in sides. Their values are centered near zero and span positive and negative regions of the number line. They also have short tails. These properties make the Gaussian distribution appropriate.

Follower velocity's histograms depart quite dramatically from those of the first two variables. In Figure \ref{fig:relative_distance_histograms_and_relative_velocity_histograms} (c), one can see some Gaussian distributions and many more mixtures of Gaussians with two, three, or four modes. The global distribution corresponding to the full data set shows a clear pattern of velocities that vary around a mode at 25 meters per second, and velocities that vary around a mode at 5 meters per second. This indicates that drivers were exposed to a ratio of congested to uncongested traffic of roughly 1:3. 

With such a broad variety of data distribution shapes, it is reasonable to take a generic approach to modeling latent model parameters and to use Gaussian random variables. Parameters that are supported by only positive or only negative values can be transformed using bijections offered by the probabilistic programming language.

\subsubsection{Output Variable Distributions}
\label{subsubsec:output_variable_distributions}

\begin{figure} 
\begin{center}
\includegraphics[width=6in]{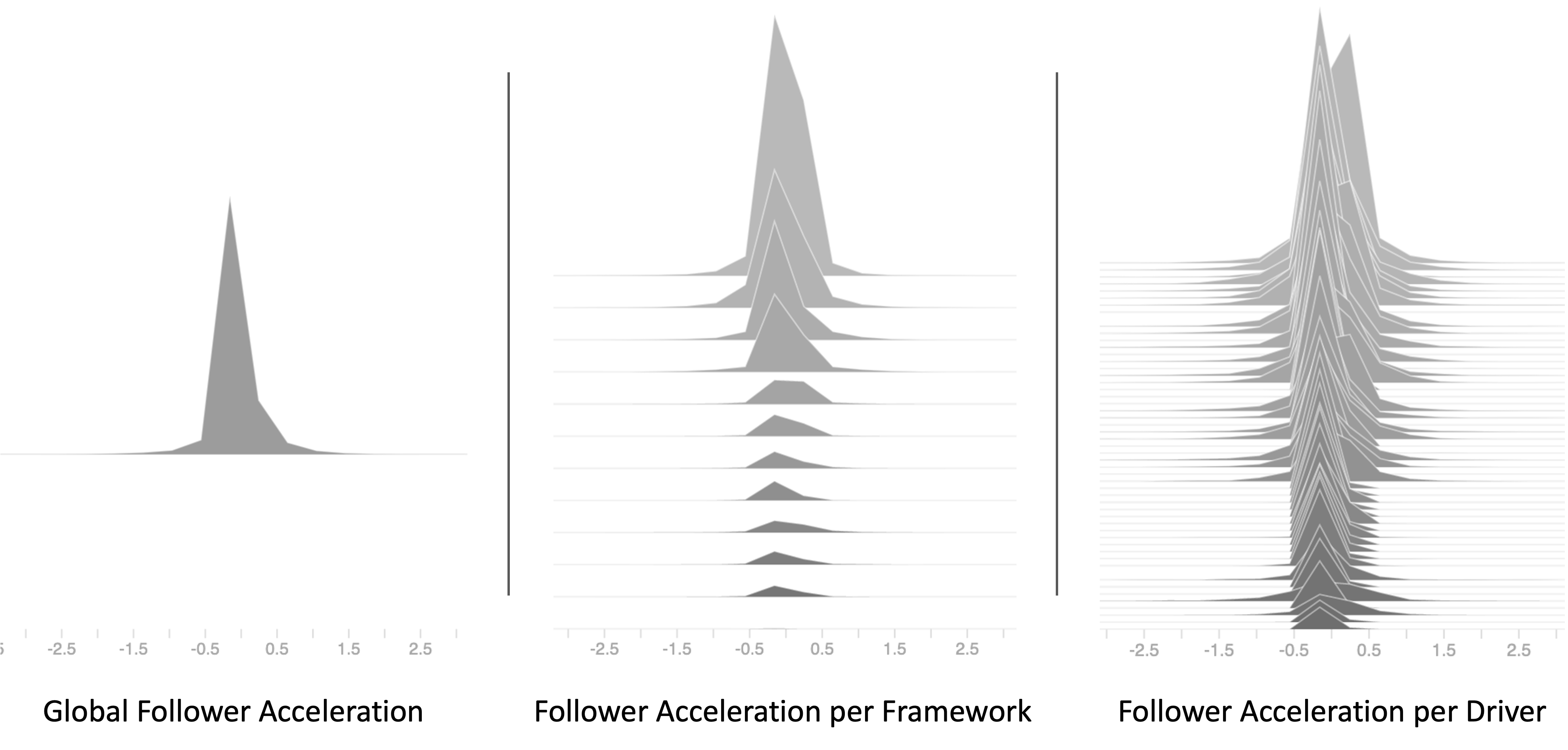}
\caption{Histograms of the output variable follower acceleration for the global population, sub-populations per framework, and sub-populations per driver.}
\label{fig:follower_acceleration_histograms}
\end{center}  
\end{figure}

The follower acceleration distributions in Figure  \ref{fig:follower_acceleration_histograms} have sharp peaks, short tails and concave-out sides. Their values are centered near zero and span positive and negative regions of the number line. The outward concavity indicates that most values are concentrated very close to zero, which is consistent with the majority of driving taking place at a cruising highway speed. These properties make the Double Gamma distribution appropriate most suitable. The Double Gamma (also known as the Reflected or Two-sided Gamma) is a flexible three-parameter distribution that ranges in shapes from a continuous spike-and-slab shape to a dumbbell shape with symmetry about a location. See Appendix \ref{appendix a}.


\subsection{Analysis of Follower Acceleration Time-series}
\label{subsubsec:analysis_of_follower_acceleration_autocorrelation}

The calibration procedure proposed in this work depends on the ability of the likelihood of the data as a whole given the latent random variables to be decomposed into probabilities of each individual data point. In turn, each data point is expected to be independent and identically distributed. Because the data are in a time-series form, one can expect the data points in a CF instance to be strongly autocorrelated and not i.i.d.. Yet, the FHWA Work Zone Driver Model enables the data to be treated as i.i.d. in part because it directly incorporates observations from past timesteps into predictions about the next timestep via its perception reaction time parameters. 

\begin{figure} 
\begin{center}  
\begin{tabular}{cc} 
\includegraphics[width=2.8in]{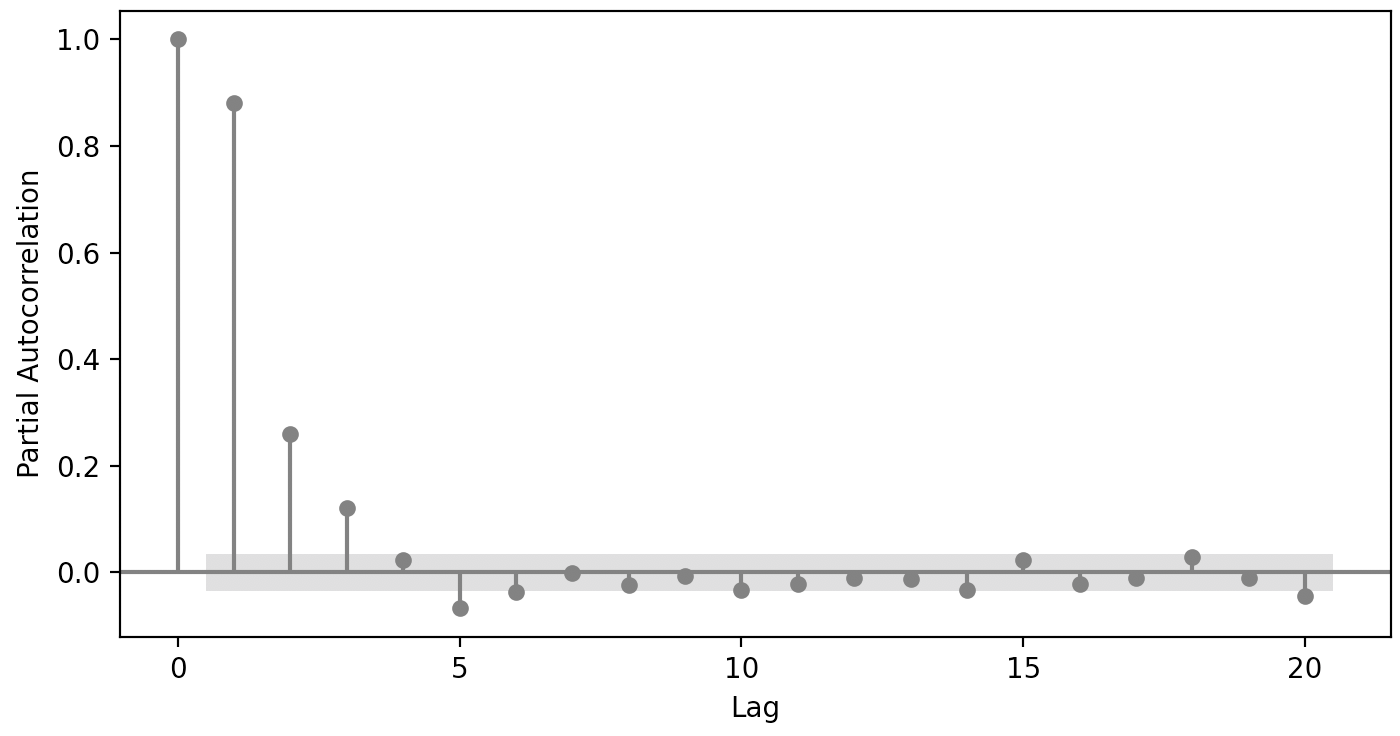} & \includegraphics[width=2.8in]{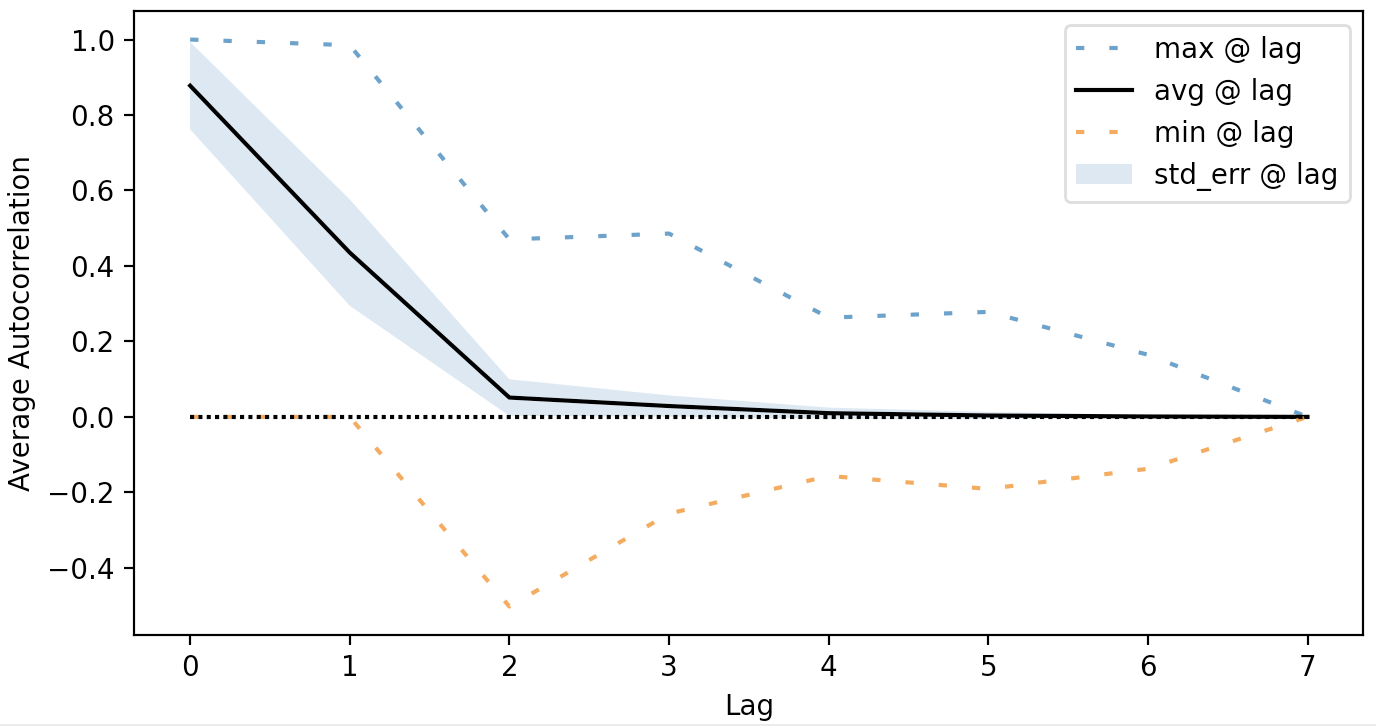} \\
(a) & (b) \\
\end{tabular}
\begin{tabular}{c} 
\includegraphics[width=2.8in]{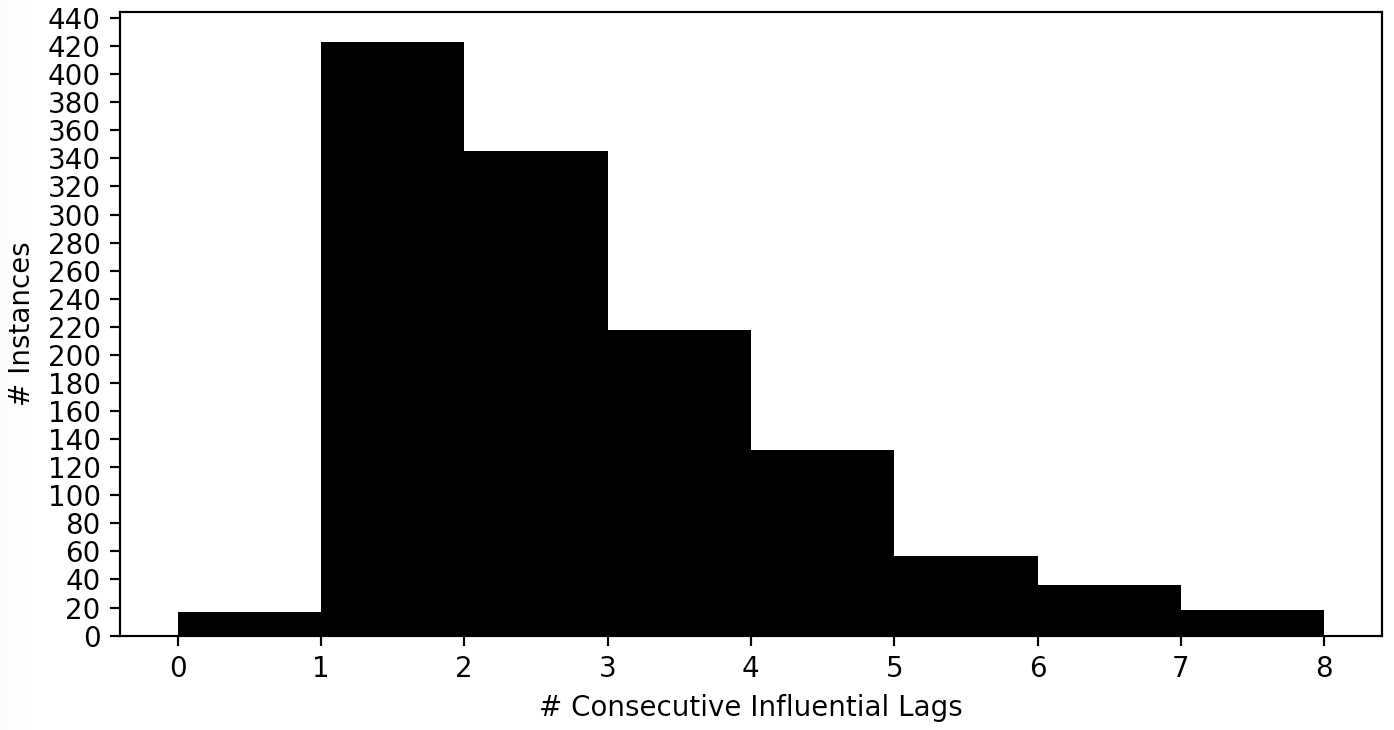}  \\
(c) \\
\end{tabular}
\caption{(a) The partial autocorrelation plot for one CF instance. In this example, the number of significant past timesteps is four. (b)  Plot of average autocorrelation per lag over all CF instances.(c) Histogram of CF instance frequency over the number of consecutive statistically influential past timesteps for each instance.}
\label{fig:autocorrelation_lag_histogram_and_autocorrelation_example_and_average_autocorrelation_plot}
\end{center}  
\end{figure}

To determine the best range over the number of past timesteps to consider using, the partial autocorrelation of the data can be studied. Partial autocorrelation measures the correlation between a variable at time $t$ and itself at some earlier time $t-k$, and for which the measure is determined by only the relationship between $t$ and $t-k$ and not any timesteps in between. This is a key differentiator between partial autocorrelation and autocorrelation, which does not exclude influences from intermediate points. Figure \ref{fig:autocorrelation_lag_histogram_and_autocorrelation_example_and_average_autocorrelation_plot} (a) shows the partial autocorrelation plot for one example CF instance. Starting at $t$ vs itself at index zero, the plot shows that the first three past timesteps should be included in a model of this series to maximize predictive accuracy. Sub-figures (b) and (c) present measures of partial autocorrelation aggregated across the entire data set. In (b) is a plot of the average of partial autocorrelation over all instances for each of the first seven sequential timesteps starting at $t-1$. In (c), on the x-axis, the number of consecutive significant past timesteps for a given CF instance, and on the y-axis is the number of instances that many significant timesteps. Sub-figures (b) and (c) reveal that the first four past timesteps are the most important to include in the model.

%% file: chap6.tex
\chapter{Methods}	\label{chapter 6}

To explore the full power of hierarchical modeling, we additionally organize car-following instances into groups per driver. Thus, every instance belongs to one framework and one driver, which has implications for how the probabilistic version of the WZDM is designed as is detailed in Chapter \ref{chapter 5}.

\section{Data Structure and Organization}

Data points within instances that belong to the same subgroup, whether that be per-framework, per-driver, or per intersection of framework and driver, are grouped into one batch per subgroup. Under the assumption that all data points in a batch are i.i.d., the log-probability of the latent parameter values conditioned on the observed data batch is equal to the sum of the log-probabilities of the individual data points. Although the data were determined to be strongly autocorrelated in Section \ref{subsubsec:analysis_of_follower_acceleration_autocorrelation}, it is acceptable to treat them as i.i.d. This is because the dynamics of longitudinal car-following are completely governed by the deterministic CF model, including the influence on past timesteps, in addition to the current timestep, on the prediction of the next time step. Thus, a single set of parameters can be applied across all timesteps, as is intended in the original deterministic formulation of the WZDM, making the ordering of timesteps completely irrelevant and eliminating temporal dependencies. A probabilistic model not derived from a physics- or process-based model would typically require a unique set of distributions for each time step (e.g. a Bayesian Structural Time-series model, \cite{scott}, or a Bayesian Recurrent Neural Network, \cite{fortunado}).

The batching approach can be applied to the IDM and WZDM models, but the W99 model requires a different data organization. Because the initial previous driving regime latent variable is specific to each car-following instance, the data points are batched per-instance rather than per sub-group.

\section{Random Variable Specification}

\begin{figure}
\begin{center}  
\includegraphics[scale=0.6]{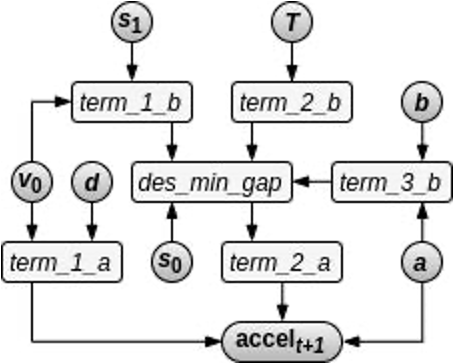}
\caption{Graph representation of the probabilistic Intelligent Driver Model. The deterministic operations of the likelihood function are abstracted into labeled rectangles, and random variables are represented by circles.}  
\label{fig:probabilistic_idm_diagram}
\end{center}  
\end{figure}

For all three car-following models explored in this project, a general framework for selecting the random variables that correspond to the model parameters was followed based on the data analysis in the preceding chapter.

The distributions of random variables in each probabilistic CF model can be categorized as either prior distributions or likelihood distributions, \cite{gelfand}. Prior distributions correspond to unobserved latent parameters of which posterior means are to be estimated. Likelihood distributions correspond to observed dependent variables. Observed independent variables are treated as inputs to the likelihood function no differently than when calibrating deterministic CF models. Since acceleration is the only response variable in each deterministic CF model, each probabilistic CF model has one scalar-variate likelihood distribution. 

\subsubsection{The Output Random Variable}

The Double Gamma distribution, \cite{dgapdf}, was chosen to model response acceleration based on the shape of the variable's histogram, which showed a substantial majority of its mass near zero with short and minimally populated tails to the left and right of zero. This overall shape was formed by the full population of data points, as well a nearly all per-framework and per-driver acceleration distributions. 

\subsubsection{Continuous Latent Random Variables}

Scalar latent parameters are modeled using Gaussian random variables with Gaussian hyperpriors in hierarchical models. For parameters supported by the non-negative reals only, transformed random variables, \cite{tfdistributions}, are created wherein the softplus function is applied to samples from the Gaussian and the evaluation of the log-probability is modified to account for the change in density of the search space entailed by the transformation. Parameters supported by non-positive reals only have Softplus applied followed by multiplication by -1.

For each parameter, ${\theta}_k$, if the support is $(-{\inf}, {\inf})$, then:

\begin{equation}
    \theta_k = \theta_{std}, \text{ where } \theta_{std} \sim Normal_k(\mu_k,\sigma_k),
\end{equation}
\newline
and if the support is $[0, {\inf})$, then:

\begin{equation}
    \theta_k = \theta_{pos} = Softplus(\theta_{std}, \beta), \text{ where } \theta_{std} \sim Normal_k(\mu_k,\sigma_k) \text{ and } \beta = 10^{-12},
\end{equation}
\newline
and if the support is $(-{\inf}, 0]$, then:

\begin{equation}
    \theta_k = \theta_{neg} = -Softplus(\theta_{std}, \beta), \text{ where } \theta_{std} \sim Normal_k(\mu_k,\sigma_k) \text{ and } \beta = 10^{-12},
\end{equation}
\newline
with:

\begin{equation}
    Softplus(x, \beta) = \frac{log(1 + e^{x\beta})}{\beta}.
\end{equation}
\newline
The transformed random distributions that yield $\theta_{neg}$ and $\theta_{neg}$ are henceforth referred to as $Normal^{pos}$ (or $N^{pos}$) and $Normal^{neg}$ (or $N^{neg}$), respectively.

\subsubsection{Categorical Latent Random Variables}

Categorical parameters are modeled using Multinomial random variables with Dirichlet priors in the hierarchical models. In the W99 model, one of the inputs at each time step is the driving regime at the previous time step. Categorical random variables support the choice of initial previous driving regime at the first step in the time-series. Since this value is unknown, a distribution over the set of possibilities is learned for each instance during calibration. There are four regimes, and thus four categories. In the WZDM, Multinomial random variables support the choice of the number of timesteps into the past that correspond to the different perception reaction time parameters. Based on the analysis in Section \ref{subsubsec:analysis_of_follower_acceleration_autocorrelation} in the previous chapter, the categories are 1, 2, 3 and 4 timesteps (e.g. tenths of a second).

\section{Model Pooling Formulations}

\begin{figure} 
\begin{center} 
 
\begin{tabular}{c} 
\includegraphics[scale=.8]{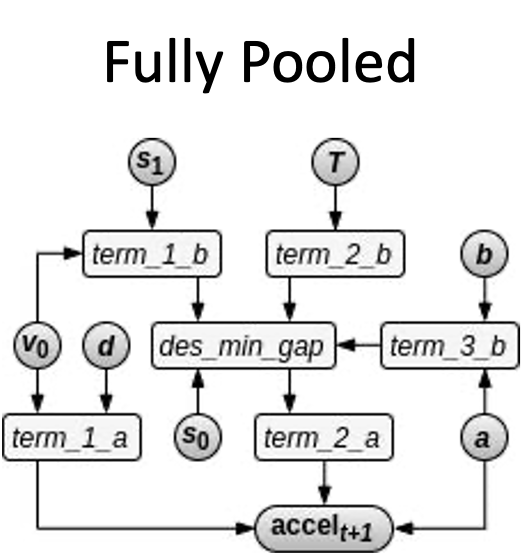}\\
\newline \\
(a) \\
\newline \\
\hline
\newline \\
\includegraphics[scale=.8]{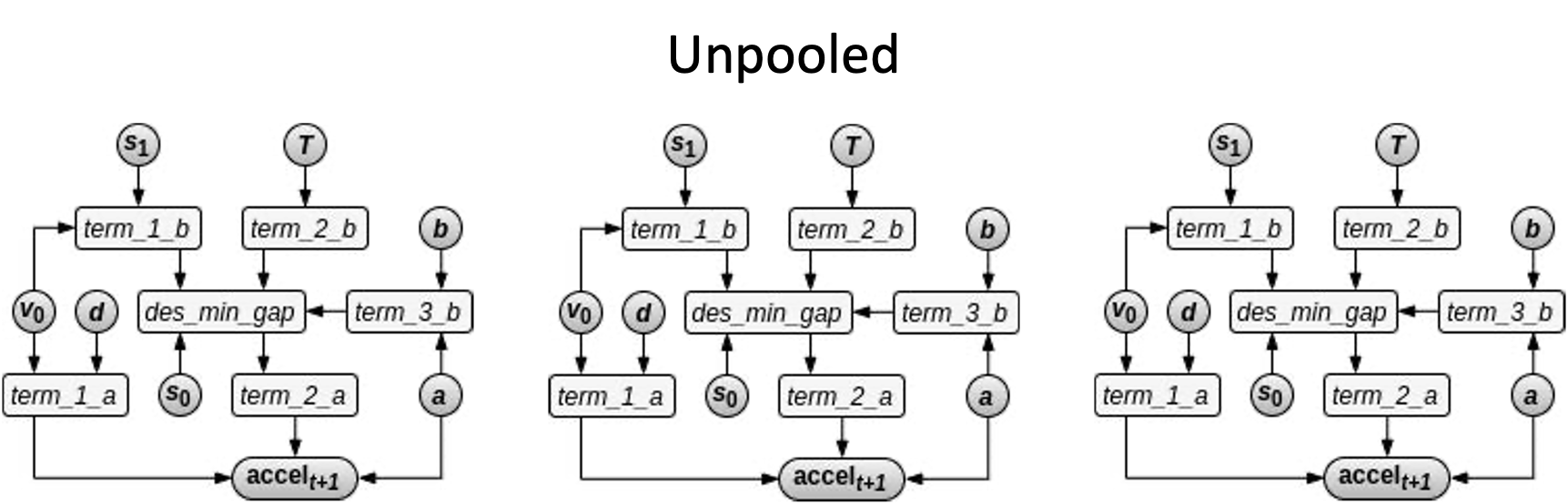}\\
\newline \\
(b) \\
\newline \\
\hline
\newline \\
\includegraphics[scale=.8]{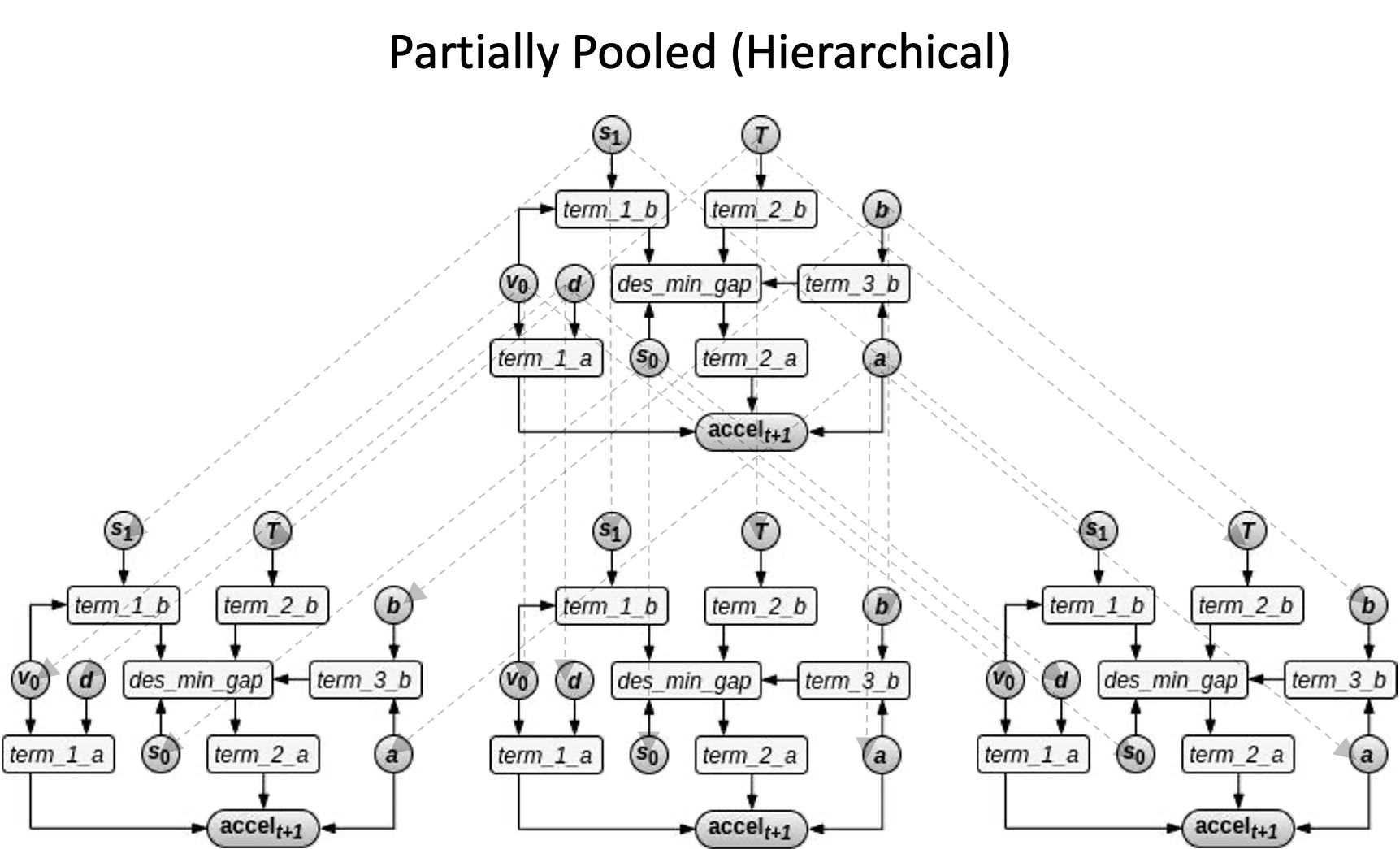}\\
\newline \\
(c) \\
\newline \\
\end{tabular}  
\caption{Histograms of the input variables relative distance (a), relative velocity (c), and follower velocity (c) for the global population, sub-populations per framework, and sub-populations per driver.}
\label{fig:pooling_formulations}
\end{center}  
\end{figure} 

Given a set of random variable assignments to model parameters, each model may be constructed with the intention of 1) estimating one set of parameters using all data, or 2) estimating one set of parameters for each subset of data per-driver or per-framework. These two cases correspond to fully pooled and unpooled probabilistic model formulations, respectively. A third formulation that is partially-pooled estimates parameters for both the full data set and each subset using a hierarchical arrangement of the parameters. Each formulation is illustrated in Figure \ref{fig:pooling_formulations} using the Intelligent Driver Model as an example, and also described below:

\begin{itemize}
	\item For a single set of parameters calibrated using data from all subsets $d \in D$, where $|D|$ equals the number of data subsets, a group of univariate-normals of dimension $k$, which is equal to the number of parameters, is used to implement a fully pooled model:
	\newline
	
    $\theta_d \sim Normal_k(\mu,\Sigma)$.
	\newline
    
    \item For $|D|$ sets of parameters, each calibrated using data from one subset $d \in D$, a set of $k$ groups of univariate normals is used to implement $|D|$ unpooled models:
	\newline
    
    $\theta_d \sim Normal_k(\mu_d,\Sigma_d)$.
	\newline
    
    \item For an alternative to either of the first two cases, a two-level hierarchy of univariate normals, for which each data subset has an separate set of $k$ random variables with their mean and standard deviation drawn from shared superset parent distributions, is used to implement a partially-pooled model:
	\newline
    
    $\theta_{d} = \mu_{d} + \sigma_{d} \: \times \: \theta_{d,norm}$;
    \newline
    $\theta_{d,norm} \sim Normal_k^{L1}(0,1)$;
    \newline
    $\mu_d \sim Normal_k^{L2}(\mu_\mu,\sigma_\mu)$;
    \newline
    $\sigma_d \sim HalfNormal_k^{L2}(\sigma_\sigma)$,
    \newline
    
    where the Half Normal distribution has non-negative support.
\end{itemize} 

\section{Probabilistic Car-following Model Construction}

\subsection{Probabilistic Intelligent Driver Model}

For the probabilistic IDM, the latent variables are $\theta$ = \{$v_0$, \textit{T}, \textit{a}, \textit{b}, $\delta$, $s_0$, $s_1$\}, the observed input variables are \textbf{x} = $\{s, v, \Delta v\}$, and the observed output variable is $\{a_F\}$. The random variable-to-parameter assignments are as follows:

\begin{center}
\scriptsize
\begin{tabular}{|l||l|l|l|l|l|l|l|}
\hline
\textbf{Params.} &  $v_0$ & \textit{T} & \textit{a} & \textit{b} & $\delta$ & $s_0$ & $s_1$ \\
\hline
\textbf{Dists.} & $N^{pos}$ & $N^{pos}$ & $N^{pos}$ & $N^{pos}$ & $N^{pos}$ & $N^{pos}$ & $N^{pos}$ \\
\hline
\textbf{Priors} & 6.5 & 1.6 & .73 & 1.67 & 4. & 2. & 0. \\
\hline
\end{tabular}
\end{center}

\subsection{Probabilistic Wiedemann '99 Model}

For the probabilistic W99, the continuous latent variables are $\theta$ = \{$CC_0$, $CC_1$, $CC_2$, $CC_3$, $CC_4$, $CC_5$, $CC_6$, $CC_7$, $CC_8$, $CC_9$, $v_0$\}, the observed input variables are \textbf{x} = $\{v_F, a_F, v_L, a_L, \Delta v, \Delta x\}$, and the observed output variable is $\{a_F\}$. The random variable-to-parameter assignments are as follows:

\begin{center}
\scriptsize
\begin{tabular}{|l||l|l|l|l|l|l|l|l|l|l|l|l|}
\hline
\textbf{Params.} & $CC_0$ & $CC_1$& $CC_2$ & $CC_3$ & $CC_4$ & $CC_5$ & $CC_6$ & $CC_7$ & $CC_8$ & $CC_9$ & $v_0$ \\
\hline
\textbf{Dists.} & $N^{pos}$ & $N^{pos}$ & $N^{pos}$ & $N^{neg}$ & $N^{neg}$ & $N^{pos}$ & $N^{pos}$ & $N^{pos}$ & $N^{pos}$ & $N^{pos}$ & $N^{pos}$ \\
\hline
\textbf{Priors} & 1.5 & 1.3 & 4. & -12. & -.25 & .35 & .0006 & .25 & 2. & 1.5 & 33.3 \\
\hline
\end{tabular}
\end{center}

The categorical variable, \textit{initial\_previous\_regime}, which is not part of the standard definition of the model, is drawn from a \textit{Multinomial} with the probs argument initialized to [.25, .25, .25, .25], indicating no prior knowledge of the distribution. Each car-following instance is assigned its own variable.

\subsection{Probabilistic FHWA Work Zone Driver Model}

For the probabilistic WZDM, the latent variables are organized into two groups; one for framework parameters and another for driving regime parameters. The observed input variables are \textbf{x} = $\{v_F, a_F, v_L, a_L, \Delta v, \Delta x\}$, and the observed output variable is $\{a_F\}$.

The continuous regime parameters are $\theta_{regime}$ = \{$G_{max}$, $G_{min}$, $G_c$, $G_s$, $V_{s,G_{max}}$, $V_{s,G_{min}}$, $V_{a,G_{max}}$, $V_{a,G_{min}}$\}. To make the search space easier to navigate, prior knowledge about how some of the regime parameters relate to others is encoded by defining new parameters. Specifically, since 
$G_s < G_c < G_{min} < G_{max}$, as is visible in Figure \ref{fig:wzdm_psychophysical_plane}, we replace the latter three parameters with $G_c\_{\Delta}G_s$, $G_{min}\_{\Delta}G_c$, and $G_{max}\_{\Delta}G_{min}$, and then when constructing regional polygons on the psychophysical plane, the original parameter values are reconstructed by adding delta values to the $G_s$ baseline value. Similarly, $V_{s,G_{min}} < V_{s,G_{max}}$, and because the $V_s$ and $V_a$ parameters are symmetric (by design) about the y-axis, $V_{a,G_{min}} = -V_{s,G_{min}}$ and $V_{a,G_{max}} = -V_{s,G_{max}}$. Thus the actual parameter vector in the implementation is $\theta_{regime}$ = \{$G_{min}\_{\Delta}G_c$, $G_{max}\_{\Delta}G_{min}$, $G_c\_{\Delta}G_s$, $G_s$, $V_{s,G_{max}}\_{\Delta}V_{s,G_{min}}$, $V_{s,G_{min}}$\}.

The random variable-to-parameter assignments are as follows:

\begin{center}
\scriptsize
\begin{tabular}{|l||l|l|l|l|l|l|l|l|l|l|l|l|}
\hline
\textbf{Params.} & $G_{max}\_{\Delta}G_{min}$ & $G_{min}\_{\Delta}G_c$ & $G_c\_{\Delta}G_s$ & $G_s$ & $V_{s,G_{max}}\_{\Delta}V_{s,G_{min}}$ & $V_{s,G_{min}}$ \\
\hline
\hline
\textbf{Prior} $\boldsymbol{\theta}$ & $N^{pos}$ & $N^{pos}$ & $N^{pos}$ & $N^{pos}$ & $N^{pos}$ & $N^{pos}$ \\
\hline
\textbf{FW1 Priors} & 28. & 4. & 5. & 3. & 0. & 2. \\
\hline
\textbf{FW2 Priors} & 55. & 5. & 7. & 3. & 0. & 2. \\
\hline
\textbf{FW3 Priors} & 30. & 4. & 5. & 3. & 0. & 2. \\
\hline
\textbf{FW4 Priors} & 25. & 5. & 4. & 6. & 3. & 2. \\
\hline
\textbf{FW5 Priors} & 30. & 4. & 5. & 3. & 0. & 2. \\
\hline
\textbf{FW6 Priors} & 25. & 5. & 4. & 6. & 3. & 2. \\
\hline
\textbf{FW7 Priors} & 30. & 4. & 5. & 3. & 0. & 2. \\
\hline
\textbf{FW8 Priors} & 25. & 5. & 4. & 6. & 3. & 2. \\
\hline
\textbf{FW9 Priors} & 30. & 4. & 5. & 3. & 0. & 2. \\
\hline
\textbf{FW10 Priors} & 25. & 5. & 4. & 6. & 3. & 2. \\
\hline
\textbf{FW11 Priors} & 30. & 4. & 5. & 3. & 0. & 2. \\
\hline
\textbf{FW12 Priors} & 25. & 5. & 4. & 6. & 3. & 2. \\
\hline
\hline
\textbf{Hyperprior} $\boldsymbol{\theta}$ & $N^{pos}$ & $N^{pos}$ & $N^{pos}$ & $N^{pos}$ & $N^{pos}$ & $N^{pos}$ \\
\hline
\textbf{Hyperpriors} & 29.83 & 4.5 & 4.75 & 4.25 & 1.25 & 2. \\
\hline
\end{tabular}
\end{center}

The continuous framework parameters are $\theta_{fw_{cont}}$ = \{$C_v$, $C_{des}$, $C_{prox}$, $C_{gap}$, $T_{safe}$, $C_{BL}$, $V_{des}$, $A_{max}$, $D_{max}$, $D_{emr}$\}. Again, prior knowledge about the relationship between $D_{max}$ and $D_{emr}$, specifically that $D_{max} < D_{emr}$, is encoded by creating a new parameter $D_{emr}\_{\Delta}D_{max}$. Thus the actual parameter vector in the implementation is $\theta_{fw_{cont}}$ = \{$C_v$, $C_{des}$, $C_{prox}$, $C_{gap}$, $T_{safe}$, $C_{BL}$, $V_{des}$, $A_{max}$, $D_{max}$, $D_{emr}\_{\Delta}D_{max}$\}. For the purpose of compactness, the table below uses $D_{emr}$ to refer to $D_{emr}\_{\Delta}D_{max}$.

The random variable-to-parameter assignments are as follows:

\begin{center}
\scriptsize
\begin{tabular}{|l||l|l|l|l|l|l|l|l|l|l|l|l|}
\hline
\textbf{Params.} & $C_v$ & $C_{des}$ & $C_{prox}$ & $C_{gap}$ & $T_{safe}$ & $C_{BL}$ & $V_{des}$ & $A_{max}$ & $D_{max}$ & $D_{emr}$\\
\hline
\hline
\textbf{Prior} $\boldsymbol{\theta}$ & $N^{pos}$ & $N^{pos}$ & $N^{pos}$ & $N^{pos}$ & $N^{pos}$ & $N^{pos}$ & $N^{pos}$ & $N^{pos}$ & $N^{neg}$ & $N^{neg}$ \\
\hline
\textbf{FW1 Priors} & 8. & .25 & 8. & 5. & .85 & 5. & 20. & 4. & -2. & -1 \\
\hline
\textbf{FW2 Priors} & 10. & .25 & 8. & 5. & 2. & 10. & 30. & 4. & -2. & -1 \\
\hline
\textbf{FW3 Priors} & 8. & .25 & 8. & 5. & .85 & 5. & 20. & 4. & -2. & -1 \\
\hline
\textbf{FW4 Priors} & 15. & .5 & 8. & 5. & 2. & 10. & 30. & 4. & -2. & -1 \\
\hline
\textbf{FW5 Priors} & 8. & .25 & 8. & 5. & .85 & 5. & 20. & 4. & -2. & -1 \\
\hline
\textbf{FW6 Priors} & 15. & .5 & 8. & 5. & 2. & 10. & 30. & 4. & -2. & -1 \\
\hline
\textbf{FW7 Priors} & 8. & .25 & 8. & 5. & .85 & 5. & 20. & 4. & -2. & -1 \\
\hline
\textbf{FW8 Priors} & 15. & .5 & 8. & 5. & 2. & 10. & 30. & 4. & -2. & -1 \\
\hline
\textbf{FW9 Priors} & 8. & .25 & 8. & 5. & .85 & 5. & 20. & 4. & -2. & -1 \\
\hline
\textbf{FW10 Priors} & 15. & .5 & 8. & 5. & 2. & 10. & 30. & 4. & -2. & -1 \\
\hline
\textbf{FW11 Priors} & 8. & .25 & 8. & 5. & .85 & 5. & 20. & 4. & -2. & -1 \\
\hline
\textbf{FW12 Priors} & 15. & .5 & 8. & 5. & 2. & 10. & 30. & 4. & -2. & -1 \\
\hline
\hline
\textbf{Hyperprior} $\boldsymbol{\theta}$ & $N^{pos}$ & $N^{pos}$ & $N^{pos}$ & $N^{pos}$ & $N^{pos}$ & $N^{pos}$ & $N^{pos}$ & $N^{pos}$ & $N^{neg}$ & $N^{neg}$ \\
\hline
\textbf{Hyperpriors} & 11.08 & .3542 & 8. & 5. & 1.425 & 7.5 & 25. & 4. & -2. & -1 \\
\hline
\end{tabular}
\end{center}

The categorical framework parameters are $\theta_{fw_{cat}}$ = \{$PRT_{{\Delta}v}$, $PRT_{{\Delta}x}$, $PRT_v$\}, in units of timesteps rather than seconds. 

The random variable-to-parameter assignments are as follows:

\begin{center}
\scriptsize
\begin{tabular}{|l||l|l|l|l|l|l|l|l|l|l|l|l|}
\hline
\textbf{Params.} & $PRT_{{\Delta}v}$ & $PRT_{{\Delta}x}$ & $PRT_v$ \\
\hline
\hline
\textbf{Prior} $\boldsymbol{\theta}$ & $Multinomial$ & $Multinomial$ & $Multinomial$ \\
\hline
\textbf{FW1 Priors} & [0, 0, 1, 0] & [1, 0, 0, 0] & [0, 1, 0, 0] \\
\hline
\textbf{FW2 Priors} & [0, 0, 0, 1] & [1, 0, 0, 0] & [0, 1, 0, 0] \\
\hline
\textbf{FW3 Priors} & [0, 0, 1, 0] & [1, 0, 0, 0] & [1, 0, 0, 0] \\
\hline
\textbf{FW4 Priors} & [0, 0, 0, 1] & [1, 0, 0, 0] & [0, 1, 0, 0] \\
\hline
\textbf{FW5 Priors} & [0, 0, 1, 0] & [1, 0, 0, 0] & [1, 0, 0, 0] \\
\hline
\textbf{FW6 Priors} & [0, 0, 0, 1] & [1, 0, 0, 0] & [0, 1, 0, 0] \\
\hline
\textbf{FW7 Priors} & [0, 0, 1, 0] & [1, 0, 0, 0] & [1, 0, 0, 0] \\
\hline
\textbf{FW8 Priors} & [0, 0, 0, 1] & [1, 0, 0, 0] & [0, 1, 0, 0] \\
\hline
\textbf{FW9 Priors} & [0, 0, 1, 0] & [1, 0, 0, 0] & [1, 0, 0, 0] \\
\hline
\textbf{FW10 Priors} & [0, 0, 0, 1] & [1, 0, 0, 0] & [0, 1, 0, 0] \\
\hline
\textbf{FW11 Priors} & [0, 0, 1, 0] & [1, 0, 0, 0] & [1, 0, 0, 0] \\
\hline
\textbf{FW12 Priors} & [0, 0, 0, 1] & [1, 0, 0, 0] & [0, 1, 0, 0] \\
\hline
\hline
\textbf{Hyperprior} $\boldsymbol{\theta}$ & $Dirichlet$ & $Dirichlet$ & $Dirichlet$ \\
\hline
\textbf{Hyperpriors} & [.917, .917, 4., 4.] & [11., .917, .917, .917] & [5.17., 6.83, .917, .917] \\
\hline
\end{tabular}
\end{center}

\section{Calibration Methods}

\subsection{Calibration using Bayesian Inference}

\subsubsection{Choosing an MCMC Algorithm}

Two MCMC algorithms are used in this work, Random Walk Metropolis-Hastings (RWMH) and Hamiltonian Monte Carlo (HMC). RWMH is introduced in Chapter \ref{chapter 3}. HMC is an advanced algorithm that uses gradients to inform its proposals for the next state to be explored during inference, \cite{hmc}. HMC can require many fewer iterations to converge than alternative algorithms like RWMH. To calibrate the probabilistic IDM, HMC is used because the model equations are differentiable. For W99 and WZDM models, RWMH is used because it does not require models to be differentiable, and these models have discrete sub-components that are conditionally selected for use at each timestep.

\subsubsection{Testing for MCMC Convergence}

To determine at what number of burn-in steps that the search for the target joint distribution converges, the searches are run multiple times with each successive run including an additional $2500$ steps, starting with $2500$ at the base run. The search is considered to have converged on a solution when the difference in joint probability between two consecutive runs falls below an arbitrary threshold.

\subsection{Calibration using Evolutionary Optimization}

To discover the best solution achievable using the differential evolution algorithm, an appropriate fitness function must be used, and the best possible set of hyperparameter values must be identified given the model and data. Two search methods are used in this work, grid search and Bayesian Optimization.

\subsubsection{Choosing a Fitness Function}

The fitness function used is the average of the root mean squared errors per car-following instance or sub-group batch, plus an additional regularization term of the Euclidean distance between the candidate parameter values and the prior values given in the literature. The magnitude of the regularization term's contribution to the model equations is scaled by an additional hyperparameter, lambda. 

\subsubsection{Hyperparameter Tuning via Grid Search}

Grid search exhaustively tests a fixed set of combinations of hyperparameter values, performing calibration once for each combination. Crossover probabilities between 0.1 and 0.9 with a step size of 0.2, differential weights between 0.1 and 1.9 with a step size of 0.2, and lambda between 0 and 0.0001 with step sizes of 0.0000025 were considered. 

\subsubsection{Hyperparameter Tuning via Bayesian Optimization}

Bayesian Optimization (BO), \cite{BayesOpt}, is one variant of sequential model-based optimization (SBMO), a class of algorithms that search a function space for functions with optimal outputs given some particular inputs. In each iteration of a search, SBMOs measure the utility of points in the parameter space and choose the point thought to yield the best output of a function in the function space evaluated on those parameters. Bayesian optimization can be considered to improve on grid search because it can require fewer function evaluations and because it can discover parameter values that fall inside the gaps in continuous space that would go unevaluated in a grid search.

%% file: chap7.tex
\chapter{Experimental Results}   \label{chapter 7}


\section{Baseline Experiment using the Intelligent Driver Model}

In this experiment, the Intelligent Driver Model was used to establish the ability of the Bayesian framework to address the research questions outlined in Chapter \ref{chapter 1} before proceeding with the development of the much more complex Wiedemann '99 model and FHWA Work Zone Driver Model. The IDM has only three observed variables and seven parameters, compared to six and 11 for W99, and to six and 18 for WZDM. Further, its formulas are simpler and easier to interpret. To develop the calibration framework quickly, this experiment used a subset of the data set used in the other experiments that only contained 209 instances, but still included all 53 drivers. The results of calibration and their consequences for the five research questions listed below are discussed.

\begin{enumerate}
	\item In general, can a calibration procedure based on Bayesian inference produce measurably better results than one based on a genetic algorithm when available field data is low in quantity? 
	\item Given the use of Bayesian calibration, how do partially-pooled models that are structured hierarchically improve calibration results over fully pooled models when simulating the full population of car-following instances, if at all?
	\item How do the partially-pooled models improve on unpooled models when simulating per-framework or per-driver car-following instances, if at all? 
    \item Can a Bayesian calibration procedure provide a rigorous way to measure the sufficiency of the size of a data set?
    \item Can the Bayesian approach to model validation give a more convincing measure of a model's goodness of fit than the hypothesis tests traditionally used in car-following model calibration?
\end{enumerate} 

\subsection{Addressing Research Question 2}

\begin{table}
\caption{Fully Pooled IDM Calibration Results}
\label{tab:fully_pooled_idm_calibration_results}
\begin{center}
\begin{tabular}{|l||l|l|l|l|l|}
\hline
\textbf{Prior} $\boldsymbol{\sigma}$\textbf{:}  & -- & .5 & 1. & 10. & 100. \\
\hline
\hline
\textbf{Param.} & \textbf{Prior} $\boldsymbol{\mu}$ & \multicolumn{4}{c|}{\textbf{Posterior} $\boldsymbol{\mu}$} \\
\hline
$v_0$ & 33.33 & 33.358 & 1.1768 & 11.176 & 81.642 \\
\hline
\textit{T} & 1.6 & 0. & .2491 & 1.1198 & 9.7124 \\
\hline
\textit{a} & .73 & .0014 & 0. & 0. & 0.  \\
\hline
\textit{b} & 1.67 & 207.27 & 292.55 & 921.98 & 2891.3 \\
\hline
$\delta$ & 4. & 3.7497 & 1.0231 & 6.5504 & 66.213 \\
\hline
$s_0$ & 2. & .0008 & .9094 & 5.1712 & 28.397 \\
\hline
$s_1$ & 0. & .001 & .5916 & 4.3731 & 41.215 \\
\hline
\hline
\textbf{RMSE} & 13.068 & .1971 & .1834 & .1593 & .1529 \\
\hline
\end{tabular}
\end{center}
\end{table}

A discussion about Question 2 will set the stage for a discussion about Question 1, and so that is done here first. Question 2 can be addressed by comparing the entries in Table \ref{tab:fully_pooled_idm_calibration_results} and Table \ref{tab:partially_pooled_l2_idm_calibration_results}, and observing the effect that the two forms of regularization have on the parameter estimates. Recall that the strength of the influence of the prior, which is governed by the value of $\sigma$, has a regularizing effect on the search, with regularization increasing as $\sigma$ decreases, and that use of a hierarchical arrangement of parameters also induces regularization, both for the super-population parameters and the sub-population parameters. The results show that combining a strongly informative prior with a hierarchical model structure yields the only results that are physically realistic while still improving on the evaluation metric, Root Mean Squared Error (RMSE) relative to the default values.

 Table \ref{tab:fully_pooled_idm_calibration_results} shows results for the fully pooled formulation of IDM, with one calibration performed for each of four choices of prior strength. Generally, the posterior mean estimates deviate from the prior mean values the least when $\sigma$ is at its lowest value, and the most when $\sigma$ is at its highest value. The root mean squared errors are negatively correlated with regularization, increasing monotonically as the initial value of $\sigma$ decreases. The means and their RMSE are compared rather than the full posterior distributions because the probabilistic models are also later compared with a model calibrated using Differential Evolution, which produces point estimates.

Table \ref{tab:partially_pooled_l2_idm_calibration_results} shows results for the super-population (level-2) parameters of the partially-pooled IDM, using the same options for prior $\sigma$. The regularization trend continues as evidenced by the RMSE values. Estimates start to fall within the ranges given in [IDM paper] when $\sigma$ equals 1, but it is only when $\sigma$ equals .5 that that all values are consistent with this particular data set. Specifically, the value of $b$, which is the "comfortable deceleration" value falls for the first time below the value that is known to be the maximum acceleration of the instrumented research vehicle used to collect the data, which is just over $4 m/s^2$. As the true maximum acceleration, this is a generous upper bound for the maximum deceleration, which is certainly lower.

\begin{table}
\caption{partially-pooled (L2) IDM Calibration Results}
\label{tab:partially_pooled_l2_idm_calibration_results}
\begin{center}
\begin{tabular}{|l||l|l|l|l|l|}
\hline
\textbf{Prior} $\boldsymbol{\sigma}$\textbf{:} & -- & .5 & 1. & 10. & 100. \\
\hline
\hline
\textbf{Param.} & \textbf{Prior} $\boldsymbol{\mu}$ & \multicolumn{4}{c|}{\textbf{Posterior} $\boldsymbol{\mu}$} \\
\hline
$v_0$ & 33.33 & 30.413 & 27.013 & 2.7433 & .2553 \\
\hline
\textit{T} & 1.6 & .2983 & .0012 & .0024 & .0016 \\
\hline
\textit{a} & .73 & .0727 & .049 & .0083 & .0010  \\
\hline
\textit{b} & 1.67 & 3.5683 & 4.9244 & 20.819 & 38.673 \\
\hline
$\delta$ & 4. & .0134 & .0004 & .0002 & .0004 \\
\hline
$s_0$ & 2. & 2.1689& 3.5817 & .0626 & .0233 \\
\hline
$s_1$ & 0. & .6790 & .0849 & 3.0907 & 2.5064 \\
\hline
\hline
\textbf{RMSE:} & 13.068 & 3.3696 & 2.3701 & .6797 & .4361 \\
\hline
\end{tabular}
\end{center}
\end{table}

\subsection{Addressing Research Question 3}

With the benefit of the two forms of regularization clearly demonstrated in the case where a single set of parameters is desired to be shared across all sub-groups, which is all drivers in the case of IDM, the benefit to estimating sub-group parameters can be discussed. In this case, the results are inconclusive and Question 3 remains unanswered by this experiment. The results can be described as random, with some sub-group posterior means being pulled moderately toward the population mean relative to the corresponding unpooled estimates, and others pushed in the opposite direction substantially. While some partially-pooled parameters, like desired velocity, yielded reasonable values for all drivers under the hierarchical model, other parameters, like comfortable deceleration, had values ranging between a plausible one order of magnitude and an absurd four orders of magnitude. The RMSEs averaged across all drivers differed by only .0337 between the best-performing partially-pooled and unpooled models, a marginal difference. In fact, the partially-pooled RMSE was the higher of the two, which is inconsistent with expected regularization behavior.

\subsection{Addressing Research Question 1}

\begin{table}
\caption{Differential Evolution IDM Calibration Results}
\label{tab:de_idm_calibration_results}
\begin{center}
\begin{tabular}{|l||l|l|l|l|l|}
\hline
\multicolumn{2}{|l|}{\textbf{Regularization Scale} ($\boldsymbol{\alpha}$):} & .95 & .65 & .35 & .05 \\
\hline
\hline
\textbf{Param.} & \textbf{Initial State} & \multicolumn{4}{c|}{\textbf{Terminal State}} \\
\hline
$v_0$ & 33.33 & 33.306 & 33.303 & 33.302 & 33.323 \\
\hline
\textit{T} & 1.6 & .3147 & .2554 & .1783 & .049 \\
\hline
\textit{a} & .73 & .1384 & .0972 & .056 & .0126  \\
\hline
\textit{b} & 1.67 & 6.3596 & 7.4918 & 9.8503 & 24.382 \\
\hline
$\delta$ & 4. & 3.9479 & 3.9351 & 3.9215 & 3.94494 \\
\hline
$s_0$ & 2. & 1.8445 & 1.7638 & 1.5884 & .7034 \\
\hline
$s_1$ & 0. & .6186 & .5987 & 0.5566 & 0.3293 \\
\hline
\hline
\textbf{RMSE:} & 13.068 & 4.1866 & 3.4474 & 2.5214 & .9962 \\
\hline
\end{tabular}
\end{center}
\end{table}

Finally, we can return to Question 1 and compare the results of the calibration using Differential Evolution in Table \ref{tab:de_idm_calibration_results} with the partially-pooled level-2 Bayesian calibration results in \ref{tab:partially_pooled_l2_idm_calibration_results}. The method of regularization, a penalty term added to the evolutionary optimization objective function that penalizes estimates that are distant from the initial values, produces a similar trend to the Bayesian regularization, with RMSE increasing monotonically as the strength of regularization, given by the value of $\alpha$, increases. In terms of consistency with physics, the left-most two solutions in Table \ref{tab:de_idm_calibration_results} are plausible, while having RMSE values that are competitive with the hierarchical Bayesian model, but the value of comfortable deceleration, $b$, remains outside of the bounds defined by the known maximum value of the instrumented research vehicle. 

So, for this particular model, and this particular data, and the two competing calibration procedures used, it can be said that the Bayesian procedure gives a better, more useful result than the evolutionary algorithm-based procedure.

\subsection{Addressing Research Question 4}

Regarding the measurement of data size sufficiency, the influence of the regularizing prior specification on the calibration results absolutely indicates that a substantially larger data set could benefit the project. Recall that the influence of the prior in Bayesian models diminishes as the amount and completeness [REF 23] of the data increase and more closely approximate the true, real-world data distribution. In this experiment, when regularization is weak, posterior estimates deviate dramatically from plausible values. This is especially evident for the desired velocity, $v_0$, and comfortable deceleration, $b$, estimates.

\subsection{Addressing Research Question 5}

Application of PSIS-LOO cross-validation, introduced in Chapter \ref{chapter 3}, proved to be ineffective for the three probabilistic models. The technique has a built-in diagnostic indication of when it cannot be relied upon as a validation measure for a given model. The Pareto distribution used in PSIS-LOO has a parameter, tail index, which is estimated. Theoretically, this parameter must be greater than zero, but in the PSIS-LOO paper, estimated below 0.7 are considered invalid. For all three models to which PSIS-LOO was applied in this experiment, the tail index was estimated to be a negative value. This means that direct cross-validation, which is highly time-consuming, would be required to validate these models, given their data and calibration procedure.

%% file: chap8.tex
\chapter{CONCLUSION}	\label{chapter 8}

The Bayesian inference paradigm enables effective parameter estimation for physics-based models to be performed in circumstances when data supporting that estimation are limited, yielding estimates that are useful when other methods based on optimization do not, as was demonstrated in this thesis using car-following models for traffic simulators particularly. The experiments presented in Chapter \ref{chapter 7} showed that the joint application of two regularization techniques specific to the Bayesian framework, partial pooling and strongly-regularizing priors, could produce estimates for all parameters of the Intelligent Driver Model that were both within the theoretical bounds defined in the paper that introduced the model, and also consistent with the physical constraints of the instrumented research vehicle used to collect the car-following data used to fit the model. An alternative estimation approach based on the evolutionary optimization algorithm Differential Evolution, paired with a standard penalty term regularization method, did not manage to produce competitive results to the Bayesian method, even when advanced hyper-parameter tuning techniques were employed. 

\section{Challenges and Future Work}

The success of the Bayesian model calibration procedure was demonstrated for the use case wherein a single set of parameters is desired to represent the behavior of many drivers under many different driving regimes and roadway types. When applied to the use case wherein different sets of parameters are desired for each driver or driving condition, the experiments did not yield conclusive results. 

The original objective of this project was to ultimately design, calibrate and validate a probabilistic version of the Federal Highway Administration's Work Zone Driver Model using a procedure developed using the IDM model. The WZDM has built into it a natural hierarchical structure, with twelve instances of the same set of model equations having twelve instances of the same set of free parameters to estimate. This mirrors the second use case for which the experiments using IDM yielded inconclusive results. Thus, while the data analysis presented in Chapter \ref{chapter 5} and the probabilistic formulations of the WZDM in Chapter \ref{chapter 6} set the stage for a Bayesian calibration procedure to be performed on that model, more work is required to identify and understand the reason why the experiment outcomes were not consistent with expectations based on the literature review performed on Bayesian methods, which is summarized across Chapters \ref{chapter 2} and \ref{chapter 3}.

Another task for future work is the addressing of the fifth research question outlined in the introduction of this thesis. The Pareto Smoothed Importance Sampling-based Leave-one-out (PSIS-LOO) cross-validation, \cite{Vehtari2017}, proposed at the outset of this project turned out to yield reliable estimates of LOO-CV, meaning the traditional cross-validation must be performed to validate the models explored in the experiments.

\section{Source Code Availability}

The methods and results of this are intended to be reproducible, and to that end source code will be available on GitHub at the URL: \url{https://github.com/foabodo/pwie} when the repository is made public at a future date.

%% file: epilogue.tex

\bibliographystyle{alpha}
\bibliography{thesis}

%% file: appendix.tex
\appendix
\renewcommand*\appendixpagename{\normalsize \normalfont APPENDICES}
\renewcommand*\appendixtocname{APPENDICES}
\appendixpage
\addappheadtotoc
\noappendicestocpagenum

\begin{appendices}

\addtocontents{toc}{\setcounter{tocdepth}{-1}}

\chapter{The Double Gamma Distribution}	\label{appendix a}

The Double Gamma (a.k.a. Reflected Gamma or Two-sided Gamma) distribution has three parameters, $\mu$, $\beta$, and $\gamma$, which correspond to the location, scale, and concentration (a.k.a. shape). In an alternative parameterization that utilizes a rate parameter, typically also named $\beta$, that is equal to the inverse of the scale. In this appendix, the formulas required to compute the PDF, Shannon entropy, and other statistics are provided.

The Double Gamma distribution was newly implemented in TensorFlow Probability in support of this project using the equations.

\section{The Probability Density}

The probability density function is given in \cite{dgapdf} to be:

\begin{equation}
    f(x) = \frac{z^{\gamma - 1}e^{-z}}{2\beta\Gamma(\gamma)},
\end{equation}
with:
\begin{equation}
    z = \frac{|x - \mu|}{\beta},
\end{equation}

and where $\Gamma$ is the Gamma function.

\section{About the Mean, Median, and Mode}

As a location-scale distribution, the mean of the Double Gamma is equal to the value of its location parameter. The median and mode are also equal to the location.

\section{The Standard Deviation}

The standard deviation as a function of distribution parameters is derived from the coefficients of variation given in \cite{shin} to be:

\begin{equation}
    \sigma = \sqrt{\Gamma(\gamma)\Gamma(\gamma + 2) - \Gamma(\gamma + 1)^2},
\end{equation}

where $\Gamma$ is the Gamma function.

\section{The Shannon Entropy}

The Shannon entropy is given in \cite{Nadarajah} to be:

\begin{equation}
    H_{Sh} = log(2\beta) - (\gamma - 1)\Psi(\gamma) + log(\Gamma(\gamma)) + \gamma,
\end{equation}
with:
\begin{equation}
    \Psi(x) = \frac{d}{dx}ln\Gamma(x) = \frac{\Gamma'(x)}{\Gamma(x)}
\end{equation}

being the Digamma function, and where $\Gamma$ is the Gamma function.

\section{About the Kullback-Leibler Divergence}

While there exists a closed-form solution to the KL-divergence, \cite{kl}, between two Laplace distributions, Laplace being a special case of the Double Gamma where the shape parameter $\gamma$ equals 1, no KL-divergence between two Double Gamma distributions could be identified in the literature, and none was derived in this project. For the purposes of this project, a generic KL-divergence formula applied to samples was sufficient.

\end{appendices}

\addtocontents{toc}{\setcounter{tocdepth}{2}}